\documentclass[sigconf, nonacm]{acmart}

\usepackage{threeparttable}
\usepackage{url}
\usepackage{amsmath,amsfonts}

\usepackage[ruled,linesnumbered, vlined]{algorithm2e}
\usepackage{algorithmic}
\usepackage{graphicx}
\usepackage{booktabs}
\usepackage{textcomp}
\usepackage{mathrsfs}
\usepackage{blindtext}
\usepackage{amsthm}
\usepackage{multirow}
\usepackage{bigstrut}
\usepackage{array}
\usepackage{amsthm}
\usepackage{tabularx}
\usepackage{subfig}
\usepackage{enumitem}
\usepackage{multirow}

\usepackage{mathrsfs}
\usepackage{verbatim}
\usepackage{color}
\usepackage{tabularx}

\usepackage[normalem]{ulem}
\newtheorem{myDef}{Definition}

\newtheorem{theorem}{Theorem}
\newenvironment{myproof}[1][\proofname]{%
	\begin{proof}[#1] % 使用 proof 环境，#1 是可选参数
	}{%
	\end{proof}
}

\usepackage[nice]{nicefrac}
\usepackage{diagbox}

\usepackage[nice]{nicefrac}
\usepackage{diagbox}

\usepackage{bm}

\newcommand\vldbdoi{XX.XX/XXX.XX}
\newcommand\vldbpages{XXX-XXX}
\newcommand\vldbvolume{14}
\newcommand\vldbissue{1}
\newcommand\vldbyear{2020}
\newcommand\vldbauthors{\authors}
\newcommand\vldbtitle{\shorttitle} 
\newcommand\vldbavailabilityurl{URL_TO_YOUR_ARTIFACTS}
\newcommand\vldbpagestyle{plain}

\def\BibTeX{{\rm B\kern-.05em{\sc i\kern-.025em b}\kern-.08em
		T\kern-.1667em\lower.7ex\hbox{E}\kern-.125emX}}

\definecolor{mygray}{gray}{0.6}

\begin{document}
\title{Cardinality Estimation on Hyper-relational Knowledge Graphs}

\author{Fei TENG}
\affiliation{%
	\institution{HKUST}
	\city{Hong Kong SAR}
	\country{China}
}
\email{fteng@connect.ust.hk}

\author{Haoyang LI}
\affiliation{%
	\institution{
	PolyU}
	\city{Hong Kong SAR}
	\country{China}
}
\email{haoyang-comp.li@polyu.edu.hk}

\author{Shimin DI}
\affiliation{%
	\institution{HKUST}
	\city{Hong Kong SAR}
	\country{China}
}
\email{sdiaa@connect.ust.hk}

\author{Lei CHEN}
\affiliation{%
	\institution{HKUST \& HKUST (GZ)}
	\city{Guangzhou}
	\country{China}
}
\email{leichen@cse.ust.hk}
%\renewcommand{\shortauthors}{F. Lastname et al.}

%\titlespacing{\subsubsection}
%{0pt}   % 左边距
%{-1pt}   % 上边距
%{0pt} % 下边距

\begin{abstract}
Cardinality Estimation (CE) for query is to estimate the number of results without execution, which is an effective index in query optimization. 
Recently, CE for queries over knowlege graph (KGs) with triple facts has achieved great success.
To more precisely represent facts, current researchers propose hyper-relational KGs (HKGs) to represent a triple fact with qualifiers providing additional context to the fact.
However, existing CE methods, such as sampling and summary methods over KGs, perform unsatisfactorily on HKGs due to the complexity of qualifiers. Learning-based CE methods do not utilize qualifier information to learn query representation accurately, leading to poor performance.
Also, there is only one limited CE benchmark for HKG query, which is not comprehensive and only covers limited patterns. 
The lack of querysets over HKG also becomes a bottleneck to comprehensively investigate CE problems on HKGs. 
In this work, we first construct diverse and unbiased hyper-relational querysets over three popular HKGs for investigating CE.
Besides, we also propose a novel qualifier-aware graph neural network (GNN) model that effectively incorporates qualifier information and adaptively combines outputs from multiple GNN layers, to accurately predict the cardinality. 
Our experiments demonstrate that our model outperforms all state-of-the-art CE methods over three benchmarks on popular HKGs.
\end{abstract}
%\begin{IEEEkeywords}
%Cardinality Estimation, Knowledge Graph
%\end{IEEEkeywords}
\maketitle

\pagestyle{\vldbpagestyle}
\begingroup\small\noindent\raggedright\textbf{PVLDB Reference Format:}\\
\vldbauthors. \vldbtitle. PVLDB, \vldbvolume(\vldbissue): \vldbpages, \vldbyear.\\
\href{https://doi.org/\vldbdoi}{doi:\vldbdoi}
\endgroup
\begingroup
\renewcommand\thefootnote{}\footnote{\noindent
	This work is licensed under the Creative Commons BY-NC-ND 4.0 International License. Visit \url{https://creativecommons.org/licenses/by-nc-nd/4.0/} to view a copy of this license. For any use beyond those covered by this license, obtain permission by emailing \href{mailto:info@vldb.org}{info@vldb.org}. Copyright is held by the owner/author(s). Publication rights licensed to the VLDB Endowment. \\
	\raggedright Proceedings of the VLDB Endowment, Vol. \vldbvolume, No. \vldbissue\ %
	ISSN 2150-8097. \\
	\href{https://doi.org/\vldbdoi}{doi:\vldbdoi} \\
}\addtocounter{footnote}{-1}\endgroup

\ifdefempty{\vldbavailabilityurl}{}{
	\vspace{.3cm}
	\begingroup\small\noindent\raggedright\textbf{PVLDB Artifact Availability:}\\
	The source code, data, and/or other artifacts have been made available at \url{\vldbavailabilityurl}.
	\endgroup
}
\section{Introduction}
 
Hyper-relational Knowledge Graphs (HKGs)\cite{StarE, JF17K, wikipeople, wikidata} play a critical role in capturing complex relationships between entities in real-world applications. 
They have been widely used in graph databases~\cite{vldbgraph1,vldbgraph2,vldbgraph3}, 
recommendation systems~\cite{recommendation1, recommendation2, recommendation3}, 
and question-answering~\cite{qa1, qa2, qa3}.
Unlike traditional Triple-based Knowledge Graphs (KGs)~\cite{yago, dbpedia, freebase}, which represent data in the form of triples (subject, predicate, object), HKGs extend this representation by incorporating additional qualifiers for each fact. 
These qualifiers provide contextual information about the fact, 
making HKGs more expressive and suitable for diverse applications.
 For example, in knowledge bases like Wikidata and YAGO, qualifiers can capture temporal or spatial information, 
 such as the time period during which someone held a position, e.g., \texttt{(Barack Obama, President, USA, {(StartTime, 2009), {(EndTime, 2017)}})}.
 Similarly, in recommendation systems, qualifiers can include contextual information, such as user preferences during specific time periods or at particular locations.

Cardinality Estimation (CE) over  HKGs aims to estimate the number of results for a query without executing it~\cite{GNCE, LSS, neurSC, ALLEY}. 
This is a fundamental task critical to various domains, such as optimizing query execution in database systems~\cite{wanderjoin, JSUB, conggao1, conggao2} and predicting motif distributions in molecular graphs~\cite{protein, phylogenetic}.
For instance, accurate cardinality estimation enables graph database management systems to choose efficient join orders and optimize resource allocation by anticipating query result sizes, 
leading to faster query execution~\cite{veridkg}.
However, there is no investigation on  CE over complex HKGs. 
On the other hand, existing two types of CE approaches for traditional KGs, i.e., 
\textit{sampling-based}~\cite{ALLEY, fastest,IMPR, JSUB, wanderjoin} and \textit{learning-based}~\cite{LMKG,LSS,GNCE, neurSC} methods,
cannot be effectively applied to HKGs.
It is because
they fail to handle the additional constraints and intricate structures introduced by qualifiers.

Specifically, existing
	\textit{sampling-based} methods \cite{ALLEY,wanderjoin} retrieve a subset of answer subgraphs that satisfy the query via sampling technique like random walks, and then compute the probability of sampled answer subgraphs as
	the cardinality of query. 
	However,
	 the additional qualifier constraints increase sampling failure probability under limited sampling number, which leads to output a much smaller value compared to the real cardinality (underestimation problem)~\cite{ALLEY, LSS}. 
	On the other hand,
	existing \textit{learning-based} methods \cite{LSS, GNCE} are proposed to directly learn the correlation of KG data by modeling the likelihood of query and its cardinality in a supervised manner \cite{GNCE}.  
	Since the query is graph structured, graph neural networks (GNNs)~\cite{GIN,vldbgnn1,vldbgnn2,compgcn, quanming2} are commonly used  to  encode the query into a query-specific embedding to infer query's cardinality, such as LSS~\cite{LSS} and GNCE~\cite{GNCE}.
	However, they	fail to adequately consider the impact of qualifiers on the entire fact. 
	Also, the learning-based approaches design fixed-layer GNNs for simple query patterns, which restricts the receptive field of nodes and cannot handle complex query patterns, 
	such as cyclic queries.

	In addition to the limited capabilities of existing CE methods, the scarcity of current query-sets over HKGs also become a bottleneck to comprehensively investigate CE problem on HKGs. 
	There is only one labeled HKG query CE evaluation dataset for researchers, i.e., WD50K-QE~\cite{StarQE}, which is less diverse and comprehensive. 
	%{\color{blue}
		%For clarification, we do not use WD50K-NFOL~\cite{NQE} here since we follow the existing CE over KG methods scope on conjunctive graph query. As WD50K-NFOL~\cite{NQE} extends WD50K-QE~\cite{StarQE} by introducing disjunctive and negation queries, it is beyond the scope of this work.
		%}
	As shown in Table~\ref{statistics}, WD50K-QE covers limited query patterns, limited cardinality ranges,  limited fact pattern sizes, and inflexible query variants, thus causing data scarcity issue in training.
	Therefore, \textit{firstly}, 
	we propose three more diverse and comprehensive datasets to thoroughly explore the query cardinality estimation problem on HKGs.
	\textit{Secondly}, to overcome the limitations of CE methods over HKGs, we propose a qualifier-aware GNN model that directly incorporates qualifier information, including a strategy to generate underlying qualifier features using a pretrained generative model~\cite{lagnn,gnndis}. 
	{Additionally}, we extend GNN layers to enlarge the reception field to  handle complex hyper-relational query topology, and use a linear projection vector to compute weights for each layer, adaptively combining the embeddings  outputted by each layer.
	%Second, to overcome the limitations of CE methods over HKGs, 
	%we propose a qualifier-aware GNN model  to directly incorporate qualifiers information, which also includes a qualifier completion strategy to generate underlying qualifier features conditioned on the main triple via a pre-trained generative model. 
	%\textit{Thirdly}, we extend the layers of GNNs to enlarge the reception field to handle complex hyper-relational query topology. 
	%We design a linear projection vector compute the weight for each GNN layer and adaptively combine the embedding outputted by each layer based on the weight.
	\textit{Thirdly},  to alleviate data scarcity and increase model generalization, we also propose a simple yet effective query augmentation strategy to optimize our model, which generates variant queries and maintains consistent cardinality relationships.
	%augments each query and predicts consistent cardinality estimation.
	%by controlling the relative cardinality between queries to keep cardinality estimation consistency over the whole queryset.
	In summary, the contributions of this work are listed as follows:
	\begin{itemize}[leftmargin=*]
		\item We generate three diverse, comprehensive,  and unbiased hyper-relational query CE benchmarks over popular HKGs.
		%providing a comprehensive benchmark for cardinality estimation on HKGs.
		
		\item We propose a novel qualifier-aware GNN model that incorporates qualifier information and adaptively combines GNN layer outputs to handle complex HKG topology. 
		
		\item We propose a simple yet effective data augmentation strategy  to augment the training data to alleviate the data scarcity and increase model generalization.
		
		%{\color{blue}\item We propose a simple yet effective data augmentation strategy to keep cardinality estimation consistency over the whole queryset by the relative cardinalities between queries.}
		
		\item The comprehensive experiments demonstrate that our proposed model significantly outperforms state-of-the-art CE methods.
		%on the newly constructed diverse and unbiased HKG datasets.
		
	\end{itemize}

	\begin{table}[t]
		\caption{Important Notations}
		\small
		%\vspace{-1em}
		%	\resizebox{\linewidth}{!}{
			%		\renewcommand{\arraystretch}{2}
			\begin{tabular}{l|l}
				\hline
				\textbf{Notation} & \textbf{Definition} \\ \hline
				${G}(\mathcal{V},\mathcal{E},\mathcal{F})$ & HKG with nodes $\mathcal{V}$, edges $\mathcal{E}$ and facts $\mathcal{F}$         \\ \hline
				
				$f=(s,p,o,\mathcal{Q}\mathcal{F}_{f})$ & A fact  in HKG ${G}$  \\ \hline
				
				$G^{Q}(\mathcal{V}^{Q},\mathcal{E}^{Q},\mathcal{F}^{Q})$	& Query graph format of query $Q$     \\ \hline
				
				$f^q=(s^q,p^q,o^q,\mathcal{Q}\mathcal{F}_{f}^{q})$ & A query fact pattern in ${G}^{Q}$   \\ \hline
				$(qr_{i}, qe_{i})$ & A qualifier pair     \\ \hline
				
				%$(qr_{i,f}^{q},qe_{i,f}^{q})$ & A qualifier pair with relation $qr_{i,f}^{q}$ and entity $qe_{i,f}^{q}$ in $\mathcal{Q}\mathcal{F}^{q}$      \\ \hline
				$?s$, $?p$,$?o$ & A variable of  entity and relation \\ \hline
				
				%	$f=(s,p,o,\mathcal{Q}\mathcal{F}_{f})$ & A fact in $\mathcal{F}$   \\ \hline
				
				%	$\{(qr_{i,f}, qe_{i,f})\}_{i=1}^n$ & The qualifier pairs for $f$, denoting as $\mathcal{Q}\mathcal{F}_{f}$      \\ \hline
				$\mathbf{h}_{s}^{(i)}$ &  Embedding of atom $s$ at the $i$-th layer  \\ \hline
				$\zeta(\cdot)$  &  Composition function for qualifier pair  \\ \hline
				$\gamma(\cdot)$  &  Function combine qualifier to relation \\ \hline
				$\mathcal{N}^+(s)$ & Income neighbors of $s$ \\ \hline
				
				$\mathcal{N}^-(s)$ & Outcome neighbors of $s$ \\ \hline
				
				$\mathcal{M}_\phi$ & A pretrained qualifier generative model \\ \hline
				
				$\mathcal{Q}\mathcal{F}_{f,p}$ & The partial qualifier pairs \\ \hline
				
				$\mathcal{QF}_{f,l}$ & The incomplete qualifiers pairs  \\ \hline
				$|\mathcal{M}^Q|$ &   Number of homomorphic mapping of query $Q$      \\ \hline
				
				$\mathbf{\tilde{h}}^{(k)}_{\mathcal{Q}\mathcal{F}_f^q}$ & Embedding of qualifiers in $f^q$  in $k$-th layer 
				
				\\ \hline
				$\mathbf{\hat{h}}^{(k)}_{\mathcal{Q}\mathcal{F}_{f,l}^q}$ &  Incomplete qualifier embedding in $k$-th layer \\ \hline
				$Q_{add}$,$Q_{rm}$  &  Augmented  queries on $Q$    \\ \hline
				%&  Augmented queries  on $Q$ by removing an edge/qualifier \\ \hline
				${\Vert Q \Vert}_\mathcal{Q}$	&  The cardinality of query $Q$          \\ \hline
				${{\Vert \hat{Q} \Vert}_\mathcal{Q}}$ & Estimated cardinality of query $Q$           \\ \hline
				$\lambda$ &  Weight        \\ \hline
			\end{tabular}
			%= (s^q, p^q, o^q, \mathcal{Q}\mathcal{F}^q)
			%=\{(qr^q_j,qe^q_j)_{j=1}^m\}
			%}
		\label{tab:notation}
		%	\vspace{-1em}
	\end{table}

	\noindent \textbf{Overview.}
	In the following sections of this paper, we introduce the preliminary and related works in Section~\ref{sec:related}, propose diverse and comprehensive  hyper-relational queryset datasets in Section~\ref{sec:queryset}, and present our hyper-relational query encoder with simple but effective data augmentation-based training approach in Section~\ref{sec:model}. 
	Section~\ref{sec:experiments} introduces our experiments and we conclude the paper with future directions in Section~\ref{sec:conclusion}.

\section{Preliminary and Related Works}\label{sec:related}
In this section, we first introduce the Hyper-relational Knowledge Graphs (HKGs) and then introduce the cardinality estimation over HKGs. The important notations are listed in Table~\ref{tab:notation}.
%In our scenario, we assume that the HKG is complete thus no missing edges need to be considered and we do not consider join at edge/qualifiers.
\subsection{Hyper-relational Knowledge Graphs (HKGs)} \label{HKG}

%\sout{
%A HKG $\mathcal{G}$ is a directed and labeled hyper-relational graph with entity set $\mathcal{V}$ and relation set $\mathcal{E}$, containing a set of hyper-relational facts where each fact $f = (s, p, o, \{qr_i, qe_i\}_n)$ represents a hyper-relational edge labeled $(p, \{qr_i, qe_i\}_n)$ between nodes $s$ and $o$, ($n$'s range varying from 0 to $n_max$ , the maximum $n$ in $\mathcal{G}$, for different facts). 
%Here all the involved $s, p, o, qr_0, qe_0, ... qe_n$ are called atoms in HKG, $s, o, qe_0, ... qe_n \in \mathcal{V}$ and $p, qr_0, ... qr_n \in \mathcal{E}$.
%For each fact, $(s, p, o)$ is called main triple maintaining the connections for hyper-relational graph, where the qualifier set $\{qr_i, qe_i\}_n$, denoted as $\mathcal{Q}$, is attached to the main relation $p$ to provide extra information. 
%Each $\{qr_i, qe_i\} \in \{qr_i, qe_i\}_n $ is called a qualifier pair with relation $qr_i $ and entity $qe_i $ that provides a certain domain of information to the corresponding fact. 
%For instance, given $f = (LeBron\_James, team, Miami\_Heat, \{startYear, 2010; endYear, 2014\})$, the qualifier set $\{startYear, 2010; endYear, 2014\}$ provide time interval information to main triple $(LeBron\_James, team, Miami\_Heat)$ with two pairs $\{startYear, 2010\}$ and $\{endYear, 2014\}$ stating the starting boundary and ending boundary resp. }

	HKGs are crucial for capturing complex relationships between entities in the real-world applications, enabling more precise and contextual data representation, such as yago~\cite{yago} and wikidata~\cite{wikidata}.
%\footnote{\# haoyang: add citations and examples}
% biological data~\cite{biological1,biological2}, biographical data~\cite{biographical},  general knowledge~\cite{wikipeople,StarE,JF17K}, etc.
In general,
a HKG ${G}(\mathcal{V},\mathcal{E},\mathcal{F})$ is a directed and labeled hyper-relational graph that consists of an entity set $\mathcal{V}$, a relation set $\mathcal{E}$, and a hyper-relational fact set $\mathcal{F}=\{f = (s, p, o, \{(qr_i, qe_i)\}_{i=1}^n)\}$, where  $s, o, \{qe_i\}_{i=1}^n \in \mathcal{V}$ are entities and   $p, \{qr_i\}_{i=1}^n \in \mathcal{E}$ are relations.
In each fact in HKG $f=(s, p, o, \{(qr_i, qe_i)\}_{i=1}^n)$, $(s,p,o)$ is the main fact and 
the qualifiers $ \{(qr_i, qe_i)\}_{i=1}^n$ provide additional context for the fact $(s,p,o)$ with a relation $qr_i$ and a corresponding value/entity $qe_i$.
For instance, given $f =$\texttt{(Barack Obama, President, USA, {(StartTime, 2009)})}, the qualifier set \texttt{{(StartTime, 2009)}} provides time information.
%to main triple \texttt{(Barack Obama, President, USA)}.
%with two pairs \texttt{(startYear, 2010)} and \texttt{(endYear, 2014)}.
Particularly, triple KGs is a specific format of HKGs by setting $n=0$ for each fact $f$.

KG Embedding (KGE) approaches~\cite{FB15K,conve, RDF2Vec} propose to learn a low-dimensional vector for each KG atom (e.g., entity $e$ and relation $r$), which can be used in downstream tasks, such graph classification~\cite{graphcla}, node classification~\cite{nodecla}, etc.
%set for atoms representation in downstream tasks \cite{LPKGE,wwwkge1, wwwkge2,wwwkge3}. 
In general,
KGE models~\cite{FB15K,conve, RDF2Vec} are trained by finding a best projection for KG atoms in the embedding space such that maximizes the confidence scores for all facts in KG. 
%and minimizes the scores for facts not belong to KG. 
Similarly, HKGE models~\cite{shrinking,quanming1,RAM,StarE,JF17K,gran}, such as ShrinkE~\cite{shrinking} and Gran\cite{gran}, are trained via the same objective besides add an extra qualifier aggregation process to gather the qualifiers to main triple.  For example, ShrinkE applied a multi-dimensional box shrinking process to gather qualifiers to main triple for each HKG fact. Then, StarE~\cite{StarE}  maximizes the confidence score via message passing on the whole HKG where qualifier pairs are aggregated to relation embedding for each fact.  
%KGE models have different strengths based on model design. 
%KGE models \cite{shrinking,gran,hahe} that take single fact as the only input for embedding model are fit for link prediction task, which pays more attention to intra-fact atoms correlation \cite{GNCE}. 
%KGE models \cite{RDF2Vec, StarE} have a good grasp on KG structural information proved to be an effective feature to learning-based CE task. 
%{\color{red}Among existing HKGE models~\cite{shrinking,gran,hahe,StarE,RAM}, StarE \cite{StarE} is the only HKGE model that encodes the whole HKG via CompGCN \cite{compgcn}, thus suited for CE over HKG.}
More details can be found in comprehensive survey~\cite{kgesurvey,kgesurvey2}.

%\begin{figure}[t]
%	\centering
%	\includegraphics[width=0.5\columnwidth]{figs/topology}
%	\caption{
%		Topology for chain, star, tree, petal and flower}
%	\label{fig:illustrate_topology}
%\end{figure}

\subsection{Cardinality Estimation on HKG Query}
	In this section, we first introduce the queries of HKGs and summarize the query patterns from existing works. Then, we introduce the cardinality estimation for HKG queries.

\subsubsection{Hyper-relational Knowledge Graph Query and Query Pattern} \label{ssec:query_HKGs}
%\sout{
%\textbf{Hyper-relational fact pattern}
%A hyper-relational fact (HRF) pattern is used to match HRFs in HKG if at least one atom is replaced by variable (placeholders that can match arbitrary HKG node/edge), e.g. pattern $(?s, p, o)$ can match any HRFs ends with node $o$ by incoming edge $p$ like $(s_1, p, o)$ and $(s_1, p, o, \{qr_0, qe_0\})$; pattern $(s, p, ?o, \{qr_0, ?qe\})$ can match any HRFs starts with node $s$ and outgoing edge $p$ with qualifier relation $qr_0$ that $(s, p, o_{1}, \{qr_0, qe_0\})$ and $(s, p, o_2, \{qr_0, qe_1\})$ could be candidates as long as they are included in HKG.  
%}

In general, the HKG queries consist of a set of hyper-relational fact patterns. Specifically, a hyper-relational fact pattern $f^q=(s^q,p^q,o^q,\{(qr^q_j,qe^q_j)_{j=1}^m\})$ is used to match facts in HKGs, where each item (e.g., $s^q$ and $p^q$) can be a specific entity/relation (e.g., $s$ and $p$) or a variable (e.g., $?s$ and $?p$) that match arbitrary entities or relations. 
For example, the fact pattern $(?s, p, o)$ can match any fact with object $o$ with relation $p$, such as $(s_1, p, o)$ and $(s_1, p, o, (qr_0, qe_0))$. Another pattern like $(s, p, ?o, (qr_0, ?qe))$ can match any fact consisting of subject $s$ with relation $p$  and a qualifier relation $qr_0$, such as $(s, p, o_1, (qr_0, qe_0))$ and $(s, p, o_2, (qr_0, qe_1),(qr_2, qe_2))$. 
%{\color{blue}From above example, facts in HKG query shows a "partial matching" property, which increases the complexity of execution of HKG query. We formally define qualifier matching in \ref{def:qualifiermatching} and prove it as hard as set cover problem. Thus, all execution-based CE methods cannot be directly adapt to HKG query due to the effectiveness and efficiency requirements.}
%\begin{myDef}[Qualifier Matching For Hyper-relational Fact Pattern] \label{def:qualifiermatching}
%	
%Given a qualifier $(qr^q_j,qe^q_j)_{j=1}^m$ o  hyper-relational fact pattern $f^q$ and a qualifier $(qr_i,qe_i)_{i=1}^n$ of fact $f$ in HKG. The qualifier of $f^q$ can match the qualifier of $f$ if:
%
%\begin{itemize}[leftmargin=*]
%	\item For any bounded qualifier pair \((qr^q_j,qe^q_j)\), it must exist in \((qr_i,qe_i)_{i=1}^n\).
%	\item For any qualifier pair contain variable relation or node, i.e. \((?qr^q_j,qe^q_j)\), it must have a mapping to a qualifier pair \((qr_i,qe_i)\), where \(qe^q_j\) is mapped to \(qe_i\) and \(?qr^q_j\) is wildcard mapping to corresponding relation \(qr_i\).
%\end{itemize}
%\vspace{-5pt}
%\end{myDef} 
%
%\begin{theorem} \label{theorem:qualifiermatching}
%	Qualifier matching problem is NP-complete.
%\end{theorem}
%	\begin{proofsketch}
%		We prove that qualifier matching problem is NP-complete based on decision version set cover problem~\cite{setcover1,setcover2}. We give the full proof in appendix~\ref{sec:qualifiermatchingproof} due to space limit.
%	\end{proofsketch}

In this paper, following~\cite{NQE,betae,10.1145/3447548.3467375,galkin2022inductive},
%\footnote{\#haoyang: please add citations of disjunctive query paper}, 
we define an HKG query $Q$ as a conjunctive of fact patterns, i.e., $Q=\bigwedge_{i=1}^{n^Q} {f^q_i}$. The conjunctive fact patterns consist of multiple fact patterns that share common nodes, which can be matched to subgraphs in HKG.
Formally, given a graph query $Q$, we denote the subgraph constituted by the entities and relations in query graph as $G^{Q}(V^{Q},\mathcal{E}^Q, \mathcal{F}^Q)$.

%\sout{
%\textbf{Hyper-relational query} 
%Hyper-relational query (HRQ) is a directed, labeled hyper-relational graph $q$ containing a set of conjunctive HRF patterns with entity set $\mathcal{V}_q$ and relation set $\mathcal{E}_q$, where $\mathcal{V}_q - \mathcal{V} = Var_{V_q}$ and $\mathcal{E}_q - \mathcal{E} = Var_{E_q}$ represents variables in entities and relations resp. Each HRF pattern $(s_q, p_q, o_q, \mathcal{Q})$ in $q$ represents a hyper-relational edge connecting $s_q$ and $o_q$ by $(p_q, \mathcal{Q})$. 
%}
%
%\sout{
%Since HRQ $q$ is composed of patterns with variables, they can be used to retrieve answers in HKG by replacing the HKG atoms to the variables and ensuring the variable-replaced graph (candidate answer) is a subgraph in HKG.  
%Answers to $q$ are retrieved by matching $q$ to subgraphs over HKG, which is to match variables $Var_{V_q}$ and $Var_{E_q}$ to entities and relations in HKG and maintaining the connectivity requirements in $q$. A returned matching result, also refer to as an answer subgraph, is a subgraph of HKG homomorphic to $q$. Namely, this is equivalent to replace the variables in nodes, edges and qualifiers by atoms in KG and decide whether the variable-replaced $q$ is a subgraph in HKG. $\mathcal{M}$ denotes all the answer subgraphs of $q$, where the cardinality of $q$ is $|\mathcal{M}|$, the number of answer subgraphs.
%}

\begin{figure}[t]
%		\vspace{-1em}
	\centering
	\includegraphics[width=1\columnwidth]{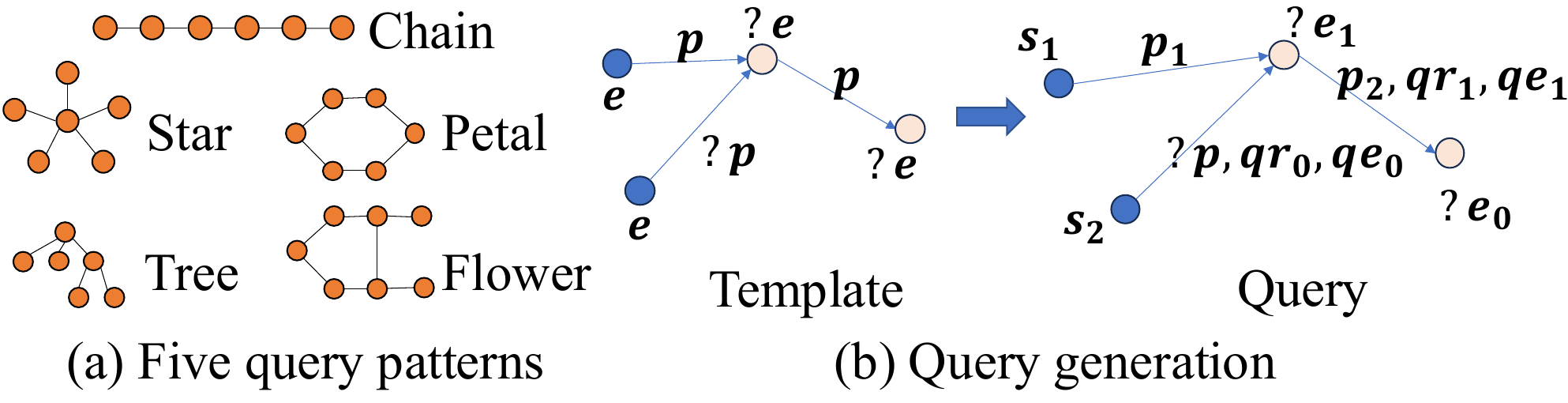}
%	\vspace{-15pt}
	\caption{Five query patterns and query generation}
	\label{fig:illustrate_topology}
%	\vspace{-15pt}
\end{figure}

{\color{black}
The queries for HKGs can form various patterns based on the connected structure of the main triples, and these patterns can summarize the real user queries from real-world applications.
 %patterns in a hyper-relational query (HRQ) form various topologies based on the connected structure of the main triples. 
 By summarizing previous work \cite{GCARE,sparqllogs} and investigating existing hyper-relational knowledge graphs (HKGs) \cite{StarE, JF17K}, the queries mainly can be summarized into five patterns, i.e., chain, tree, star, petal, and flower. 
 Figure~\ref{fig:illustrate_topology}~(a)
 %\footnote{\#haoyang: please add figures} 
 provides a clear illustration of each query pattern, where nodes refer to the subject $s$ and object $o$ in the main triple $(s,p,o)$ and  all qualifiers are omitted for simplicity, since all qualifiers as extra constraints on main facts.
 \begin{itemize}[leftmargin=10pt] 
 	\item \textbf{Chain.} Facts patterns are connected sequentially, $Q=(s_1, p1, s_2) \wedge (s_2, p2, s_3) \wedge (s_3, p3, s_4)$.
 	\item \textbf{Tree.} A topology where there is exactly one path connecting any two nodes in the queries.
 	\item \textbf{Star.} A special type of tree where exactly one node has more than two neighbors.
 	\item  \textbf{Petal.} A cyclic topology where two nodes are connected by at least two disjoint paths.
 	\item \textbf{Flower.} A topology that includes at least one petal and one chain attached to a central node .
 \end{itemize}
}

%\sout{\textbf{Query patterns} The conjunctive HRF patterns in a HRQ forms several type of topologies corresponding to the connected structure of main triple.  
%By summarizing previous work \cite{GCARE} and checking existing HKGs \cite{StarE}\cite{JF17K}, we mainly study CE problem on following five topologies, chain, tree, star, petal and flower. We provide a figure \ref{fig:illustrate_topology} for clearly illustration.
%Since we regard all qualifiers as extra constraints for edges, the term "node" in following refers to the node in main triple. 
%Chain is a topology that all nodes is connected one by one, e.g. $(s_1, p1,  s_2), (s_2, p2, s_3), (s_3, p3, s_4)$. \cite{sparqllogs} 
%Tree is that for every two nodes in HRQ there exists exactly one path connects the two nodes. \cite{sparqllogs} 
%Star is specialized tree that exists exactly one node with more than two neighbors. \cite{sparqllogs} 
%Petal is a cyclic topology that there exists two nodes connected by at least two disjoint paths. \cite{sparqllogs}
%Flower is a topology that at least one petal and one chain is attached to a central node. \cite{sparqllogs}
%}

\subsubsection{Cardinality Estimation on Hyper-relational Knowledge Graph Query}\label{ssec:ce_on_HKGs}

%\sout{
%Cardinality estimation is to estimate the number of homomorphically matched subgraphs for a hyper-relational query in HKG without execute it.  
%The accurately estimated cardinality for a query reflects the query's execution cost. 
%By knowing the approximated cardinality for sub-parts of a query in advance, the query engines can generate a cost-reduced execution order for sub-queries thus to reduce the cost for the given query.  
%Formally, given a hyper-relational query graph $q$ and a HKG $\mathcal{G}$, cardinality estimation problem is to approximately compute $|\mathcal{M}|$, the cardinality of all homomorphic embeddings of $q$ in $\mathcal{G}$.
%} 

CE refers to estimate the number of matched subgraphs for a hyper-relational query in HKG without executing it. The accurately estimated cardinality for a query can help DBMS choose efficient join orders and resource allocation and predict the frequency and distribution of specific motifs in molecular graphs.  We formally give a definition of the CE on HKG query as follows.

\begin{myDef}[Cardinality Estimation on HKG Query]\label{def:ce}
Given a query $Q$ with query graph $G^{Q}(\mathcal{V}^{Q},\mathcal{E}^Q, \mathcal{F}^Q)$,  an HKG  $\mathcal{G}(\mathcal{V},\mathcal{E}, \mathcal{F})$,
a mapping function \( m: Q \rightarrow \mathcal{G} \) that identifies a homomorphic subgraph mapping if:
	\begin{itemize}[leftmargin=10pt]
	\item For each node \( v \in \mathcal{V}^Q\), \( m(v) \in \mathcal{V} \).
	\item For each hyper-relational edge \( e = (s^q, p^q, o^q, \mathcal{Q}\mathcal{F}_{e}^q) \in \mathcal{E}^Q \), there exists an hyper-relational edge \( e' = (s, p, o, \mathcal{Q}\mathcal{F}_{e^{\prime}}) \in \mathcal{E} \) such that (1) \( m(s^q) = s \), (2)  \( m(o^q) = o \), (3) \( p^q = p \), (4) \( \mathcal{Q}\mathcal{F}_{e}^q \subseteq \mathcal{Q}\mathcal{F}_{e^{\prime}} \).
	\end{itemize}
we denote the set of all homomorphic mappings $m(\cdot)$ that match the query $Q$ in $\mathcal{G}$ as  $\mathcal{M}^{Q}$.  
The cardinality estimation on query $Q$ is to estimate the size of  the matching set  $|\mathcal{M}^Q|$ (i.e., $\Vert Q \Vert_{\mathcal{Q}}$), and it is the total number of subgraphs in $G$ that match the query graph $G^Q$.
\end{myDef}

%\sout{\textbf{Homomorphic subgraph matching} Given HRQ $q$ and HKG $\mathcal{G}$, a function $m: q \rightarrow \mathcal{G} \in M$ is called a homomorphic subgraph mapping if satisfying 1): For any $v \in \mathcal{V_q}$, $m(v) \in \mathcal{V}$, that every node in $q$ must be mapped to a node in $\mathcal{G}$ 2) for any $(v, v^{\prime}) \in \mathcal{E_q}$, $(m(v), m(v^{\prime})) \in \mathcal{E}$, which means every hyper-relational edge in $q$ must by mapped to an hyper-relational edge in HKG, besides, $\mathcal{Q}(v, v^{\prime}) \in \mathcal{Q}((m(v), m(v^{\prime}))$ which means qualifiers in every hyper-relational edge must be included in to the matched hyper-relational edge in HKG.}
%

The existing CE methods over graph data (e.g., triple KGs~\cite{GNCE,LSS,ALLEY,wanderjoin,IMPR,JSUB} and relational graphs~\cite{propertygraph1,propertygraph2,conggao1,conggao2,conggao3}) can be summarized into three classes: \textit{summary-based}, \textit{sampling-based}, and \textit{learning-based approaches}. We will  introduce them and state their limitations on HKGs.

\noindent\textbf{\underline{Summary-based Approaches.}} These works~\cite{SUMRDF, CSET, CEG} build a summary graph for the entire KGs, then execute the query on the summary graph to obtain the cardinality of queries.
However, the effectiveness of this branch is proven to be poor~\cite{GCARE}, since the unavoidable information loss in graph summary. 
%\footnote{\#haoyang: please add a reason.}

\noindent \textbf{\underline{Sampling-based Approaches.}} 
These works~\cite{ALLEY, wanderjoin,fastest} propose to use a random-walk strategy to sample several answer mapping for query in the data graph and estimate the cardinality by computing the weights for sampled answer mappings. 
However, the random-walk results in samples may not correspond to the query (sample failure), which is referred to as failed samples. 
The failed samples contribute zero weight to the final estimation thus making the estimated cardinality significantly lower than the real one.
%{\color{blue} Both summary methods and sampling methods requires the execution of HKG query. The execution requires qualifier matching either on summary graph or on random walking strategy.  According to theorem~\ref{theorem:qualifiermatching},there is an intuitive idea that to adopt existing set cover problem~\cite{setcover1,setcover2} solutions to solve qualifier matching problem when processing one fact pattern in HKG query. However, since set cover problem is NP-complete~\cite{nphard}, the exact qualifier matching can only be solved in polynomial time unless P=NP, causing prohibited time cost for CE task. If adapting an approximation algorithm, such as greedy algorithm~\cite{setcover1}, it returns a sub-optimal qualifier matching which limits the accuracy for deciding whether a fact pattern can match an HKG fact. This inaccurate qualifier matching result will reduce the matched number of HKG facts for a single fact pattern, thus affecting the CE result on whole HKG query consisting of multiple fact patterns. Thus, approaches that including query execution are infeasible in CE over HKG.}
\begin{table*} 
	\centering
	\vspace{-10pt}
	\caption{The statistics of existing dataset and our generated datasets.}
	\vspace{-10pt}
	 \setlength\tabcolsep{1pt}
	\small
%	\resizebox{\linewidth}{!}{
	%	\renewcommand{\arraystretch}{1.5}
		\begin{tabular}{c|ccccc|ccccc|cccc|cccc|cc}
			\toprule
			\multicolumn{1}{c|}{\multirow{2}{*}{}} & \multicolumn{5}{c|}{\textbf{Query Pattern}}                    & \multicolumn{5}{c|}{\textbf{Max Join Degree}}                                                                          & \multicolumn{4}{c|}{\textbf{Fact Size}}                                                                     & \multicolumn{4}{c|}{\textbf{Cardinality Range}}                                                                                                 & \multicolumn{2}{c}{\textbf{\# Bounded Nodes}}           \\ \cline {2-21}
			& \multicolumn{1}{c}{\textbf{Chain}} & \multicolumn{1}{c}{\textbf{Star}}  & \multicolumn{1}{c}{\textbf{Tree}}  & \multicolumn{1}{c}{\textbf{Petal}} & \textbf{Flower} &\multicolumn{1}{c}{\textbf{Chain}} & \multicolumn{1}{c}{\textbf{Star}}  & \multicolumn{1}{c}{\textbf{Tree}}  & \multicolumn{1}{c}{\textbf{Petal}} & \textbf{Flower} & \multicolumn{1}{c}{\textbf{\textless{}=3}} & \multicolumn{1}{c}{\textbf{6}}     & \multicolumn{1}{c}{\textbf{9}}    & \textbf{12}   & \multicolumn{1}{c}{\textbf{\textless{}$\textbf{10}^\textbf{3}$}} & \multicolumn{1}{c}{\textbf{\textless{}$\textbf{10}^\textbf{4}$}} & \multicolumn{1}{c}{\textbf{\textless{}$\textbf{10}^\textbf{5}$}} & \textbf{\textgreater{}=$\textbf{10}^\textbf{5}$} & \multicolumn{1}{c}{\textbf{0}}    & \textbf{\textgreater{}0 }  \\ \hline
			%Yago                     & \multicolumn{1}{c|}{312}   & \multicolumn{1}{c|}{318}   & \multicolumn{1}{c|}{264}   & \multicolumn{1}{c|}{321}   & 12     & \multicolumn{1}{c|}{294}           & \multicolumn{1}{c|}{416}   & \multicolumn{1}{c|}{303}  & 214  & \multicolumn{1}{c|}{640}      & \multicolumn{1}{c|}{209}        & \multicolumn{1}{c|}{189}               & 189                   & \multicolumn{1}{c|}{1227} & 0                \\ \hline
			\textbf{WD50K-QE}                 & \multicolumn{1}{c}{57517} & \multicolumn{1}{c}{19087} & \multicolumn{1}{c}{62433} & \multicolumn{1}{c}{0}     & 0   & \multicolumn{1}{c}{2} & \multicolumn{1}{c}{3} & \multicolumn{1}{c}{3} & \multicolumn{1}{c}{0}     & 0    & \multicolumn{1}{c}{139037}        & \multicolumn{1}{c}{0}     & \multicolumn{1}{c}{0}    & 0    & \multicolumn{1}{c}{139037}   & \multicolumn{1}{c}{0}          & \multicolumn{1}{c}{0}                 & 0                     & \multicolumn{1}{c}{0}    & 139037           \\ \hline
			\textbf{WD50K (ours)}             & \multicolumn{1}{c}{8800}  & \multicolumn{1}{c}{6564}  & \multicolumn{1}{c}{10284} & \multicolumn{1}{c}{1472}  & 2710   & \multicolumn{1}{c}{2}  & \multicolumn{1}{c}{12}  & \multicolumn{1}{c}{6} & \multicolumn{1}{c}{4}  & 5  & \multicolumn{1}{c}{3200}          & \multicolumn{1}{c}{13300} & \multicolumn{1}{c}{6540} & 6890 & \multicolumn{1}{c}{20246}    & \multicolumn{1}{c}{3070}       & \multicolumn{1}{c}{2491}              & 4023                  & \multicolumn{1}{c}{9830} & 20000 \\ \hline
			
			\textbf{JF17K (ours)}             & \multicolumn{1}{c}{8598}  & \multicolumn{1}{c}{2170}  & \multicolumn{1}{c}{941} & \multicolumn{1}{c}{10400}  & 2120    & \multicolumn{1}{c}{2}  & \multicolumn{1}{c}{12}  & \multicolumn{1}{c}{6} & \multicolumn{1}{c}{2}  & 5   & \multicolumn{1}{c}{2898}          & \multicolumn{1}{c}{8901} & \multicolumn{1}{c}{6540} & 1890 & \multicolumn{1}{c}{18156}    & \multicolumn{1}{c}{1869}       & \multicolumn{1}{c}{1501}              & 2703                  & \multicolumn{1}{c}{10128} & 14101 \\ \hline
			
			\textbf{Wikipeople (ours)}             & \multicolumn{1}{c}{13701}  & \multicolumn{1}{c}{2170}  & \multicolumn{1}{c}{840} & \multicolumn{1}{c}{1400}  & 2120  & \multicolumn{1}{c}{2}  & \multicolumn{1}{c}{12}  & \multicolumn{1}{c}{4} & \multicolumn{1}{c}{2}  & 5   & \multicolumn{1}{c}{7900}          & \multicolumn{1}{c}{8800} & \multicolumn{1}{c}{1641} & 1890 & \multicolumn{1}{c}{9525}    & \multicolumn{1}{c}{2360}       & \multicolumn{1}{c}{2431}              & 5915                  & \multicolumn{1}{c}{10130} & 10101  \\ \bottomrule
		\end{tabular}
	%}
	\vspace{-10pt}
	\label{statistics}
\end{table*}
\noindent\textbf{\underline{Learning-based Approaches.}} 
These approaches~\cite{LSS, GNCE} initialize the query nodes and edges by KGE. 
Then, inspired by the success of GNNs on graphs~\cite{wu2020comprehensive, wan2023scalable},
they use GNNs to learn a query representation based on query structure to predict the cardinality directly. 
However, existing learning-based methods still have a limited scope that focuses on graph data only that both initialization embedding and GNN cannot represent the qualifiers. 
Moreover, the fixed layered GNN limits the receptive field of nodes, leading to inaccuracy on several complex query patterns like petals, flowers, long chains and stars with large node degree.

\noindent \textit{\textbf{Summary}.} Compared to existing learning-based methods, we adopt StarE as initialization feature and propose an hyper-relational query encoder that injects qualifers as an extra part of hyper-relational edge in query. 
Our encoder breaks the limitation for 2-layer GNN design thus can handle multiple complex query patterns like flower and long chain by computing an adaptive weight for each GNN layer to mitigate over-smoothing problem~\cite{DEEPGNNref}.
%\footnote{\#haoyang: revise it later}

\section{Hyper-relational Queryset construction}\label{sec:queryset}

%{\color{blue}
%	+++ To Tengfei:\\
%	1. introduce exsiting stat and why exiting datasets are not good. \\
%	2. state that what are good datasets for HKGs. \\
%	3. Your method with figures, including procedure, algorithm, proof. \\
%	4. Then give what your created with stat. \\
%	+++\\
%}

In this section, we first present statistics highlighting the biased cardinality distribution and limited topologies in existing querysets for the CE problem. Then, we outline our algorithm to construct a more diverse and unbiased dataset.

\subsection{Hyper-relational Queryset} \label{querysetstat}
Based on analysis for query logs \cite{sparqllogs} and previous studies for CE problem \cite{GCARE, LSS}, we firstly identify four dimensions to generate diverse and comprehensive queryset, including 
query pattern, fact pattern size, cardinality range, and bounded node number.
% to describe a queryset. 
\vspace{-3pt}
\begin{itemize}[leftmargin=10pt]
	\item \textbf{Query Pattern.} The querysets should include a variety of query structures, i.e., chain, tree, star, petal, flower, illustrated in Figure \ref{fig:illustrate_topology} and Section~\ref{ssec:query_HKGs}.

	\item \textbf{Fact Pattern Size.}  Fact pattern size is the number of fact patterns in a hyper-relational query.
	The querysets  should  include   fact patterns ranging from simple patterns with few facts to complex patterns with many interconnected facts.
	% to evaluate performance comprehensively.

	\item \textbf{Cardinality Range.} The querysets  should  include queries with a wide range of cardinality values, from very selective queries with few results to highly general queries with many results. 
	
	\item \textbf{Bounded Node Number.} The querysets should include queries with different number of bounded nodes where specific entities are fixed to evaluate how well the methods handle constraints.
	
\end{itemize}
\vspace{-3pt}
By incorporating four dimensions, the querysets of
HKGs could provide comprehensive benchmarks, ensuring that CE approaches on HKGs are tested across a variety of realistic and challenging scenarios.
However,  there only exists one less diverse dataset on HKG query CE, i.e., WD50K-QE \cite{StarQE}. 
As show in Table~\ref{statistics}, WD50K-QE only covers limited three query patterns, limited cardinality range ($<10^3$), limited fact pattern size ($\le 3$), and inflexible bounded nodes.
Thus, there lack diverse and comprehensive datasets to explore  query cardinality estimation problem on HKGs.

\subsection{HKG Queryset  and Cardinality Generation}
In this section, we propose to generate diverse and unbiased query sets for CE over HKG with three steps, i.e., query template generation, query generation, and cardinality computation.

\noindent\textbf{\underline{Step 1: Query Template Generation}} Following~\cite{query2box,efocqa,StarQE}, we generate predefined query templates by extracting query structures from G-CARE datasets~\cite{GCARE}. The G-CARE query datasets contain various query patterns, such as flower, star, etc., and therefore we can obtain comprehensive and diverse query templates. In general, a query template is defined as a hyper-relational directed acyclic graph (HR-DAG), and the inter-connectivity (e.g., the number of nodes and edges) of each HR-DAG is fixed. As shown in Figure~\ref{fig:illustrate_topology}~(b), each node and edge in an HR-DAG is either a variable or a bounded value, which will be used to generate queries with various qualifiers, entities, etc., in Step 2.

%Thus its pattern is also fixed. Each node and edge in HR-DAG is either variable or bounded value. In step 2, we will sample entities and relations in HKG maintaining the structure of HR-DAG to materialize the bounded nodes and edges. Once all bounded nodes and edges are materialized by entities and relations in HKG, a query is generated. The generated query has the same pattern with the HR—DAG.
%To keep consistent with the queries in real-world case, we requires the query template to contain multiple variable nodes instead of only one variable compared to~\cite {query2box,StarQE}.
%A set of diverse queries with the same topology and fact size can be generated by fixing different numbers of edges and nodes. 
%We generate templates by extracting key structures and analyzing existing user queries based on \cite{GCARE}. 
%GCARE~\cite{GCARE} provided a lot of user queries for each pattern mentioned in~\ref{ssec:query_HKGs}. As each pattern has at least one template, we can directly extract the templates and perform step 2 on three HKGs to generate queries including all five patterns.

\noindent\textbf{\underline{Step 2: Query Generation}}
Given a HKG and a query template, we first generate a query by pre-order traversal. As shown in Figure~\ref{fig:illustrate_topology}~(b), we first sample an entity $s_1$ from HKG as the mapping to the root $?e$ of HR-DAG. Then, we sample a relation $p_1$ with candidate qualifiers from  edges of $s_1$. 
This process is repeated until all bounded nodes/edges in the template are filled.

\noindent\textbf{\underline{Step 3: Cardinality Computation}}
As shown in Algorithm~\ref{alg:querygeneration} in Appendix~\ref{alg:querygeneration}, we compute the cardinality via post-order traversal for the generated query. 
The post-order traversal will recursively execute on each branch of $Q$. Then merge the result at the root of each branch.
For each single branch, the algorithm starts with each leaf nodes $e_l$, we find the mapped HKG node set $V_{e_l}$ of $e_l$ and initialize an dictionary $d$ that storing $v_{e_l} \in V_{e_l}$ and current number of mappings at $v_{e_l}$. 
Then based on incident edge label $r_l$, we find the mapped HKG node set $V_{e_{l-1}}$ for $e_{l-1}$ which is the neighbor node of $e_l$. 
The algorithm updates the dictionary by replace $v_{e_l}$ to $v_{e_{l-1}}$ if $v_{e_l}$ is connected to $v_{e_{l-1}}$ by $r_l$ and add the number of mapping at $v_{e_l}$ to value of $v_{e_{l-1}}$. 
The above process will be executed until the root for the branch of $q$ is reached.
As for merging of branches joined at one node $e_i$, we intersect the nodes $v_{e_i}$ stored in each dictionary $d_i$ corresponding to branches $q_i$ first. 
Then for each intersected $v_{e_l}$, we take the product for the number of mapping in each dictionary as the joined number of mapping at $e_i$ since each branch is independent to other branches.

%{\color{black}}
%{\color{black}
\begin{theorem}
	Algorithm~\ref{alg:querygeneration} can obtain the exact cardinality for each query in $O(|\mathcal{V}|*|\mathcal{E}^Q|*|\mathcal{E}|)$.
\end{theorem}

\begin{myproof} \label{proof:querygeneration}
	Our DP-based algorithm can compute the exact cardinality for queries. Since our algorithm executes based on HR-DAG corresponding to the query, the HR-DAG maintains all nodes and edges in the acyclic query. Thus, to compute the cardinality of HR-DAG is equal to compute the cardinality for corresponding acyclic query.  Then, we will give the proof for optimality at node $e_i$. 
	
	%For each node $e_i$ in HR-DAG with one child node $e_{i+1}$, the algorithm will search the edges in HKG that connects all the candidates for $e_i$ and $e_{i+1}$. 
	%For any connected $v_{e_i}^j$ and $v_{e_{i+1}}^j$ where $v_{e_i}^j$ is a candidate node for $e_i$ and $v_{e_{i+1}}^j$ is that for $e_{i+1}$, the partial cardinality at $v_{e_i}^j$ is increased by the partial cardinality at $v_{e_{i+1}}^j$, corresponding to line $d[v_{e_(i-1)}] = d[v_{e_(i-1)}] + d[v_{e_(i)}] $. 
	%If $e_i$ has $k$ child nodes, since each sub-tree rooted at $e_{i+i^{\prime}}, i^{\prime} \in [1, k] $ is independant to the sub-trees rooted at its siblings, the parital cardinality at $v_{e_i}^j$ is $\prod_{i^{\prime}=1}^{k}|v_{e_{i+i^{\prime}}}^j|$ where $|v_{e_{i+i^{\prime}}}^j|$ is the partial cardinality at $v_{e_{i+i^{\prime}}}^j$, corresponding to line $d[v] = d[v] * d_i[v]$ in alg.~\ref{alg:querygeneration}. 
	
	The state transition equation at $e_i$ is a combination of Equation~\eqref{alg:transtion1} and Equation~\eqref{alg:transtion2}, representing the cardinality at $e_i$ ($d[e_{i}]$) is summation of cardinality at each $v$ can be mapped to $e_i$. The cardinality at $v$ ($d[v]$) is a product for cardinalities at all incident node $v^{\prime}$ that $v^{\prime}$ can be mapped to $e_{i+1}$, the child node of $e_i$. Equation~\eqref{alg:transtion1} and Equation~\eqref{alg:transtion2} covered all candidates of $e_i$ so that $d[e_{i}]$ is optimal.
	\begin{align} \label{alg:transtion1}
		d[e_{i}] = \sum_{v \rightarrow e_{i}} d[v] \\
		d[v] = \prod_{v^{\prime} \rightarrow e_{i+1}, (v, v^{\prime}) \in \mathcal{E}} d[v^{\prime}] \label{alg:transtion2}
	\end{align}
	Since the algorithm executes in bottom-up manner for each node in HR-DAG and each non-root node has exactly one ancester node, the optimality holds for every node in query by inductive hypothesis. To notice that for cyclic queries, we can convert them to HR-DAG as follows~\cite{dpdag}: We first select the node $e_c$ that forming the cycle and replicate it. Then we perform decompose the query into DAG maintaining each replicate of $e_c$ at leaf node. 
	As cyclic queries can be also convert into a HR-DAG,
	we guarantee that Algorithm~\ref{alg:querygeneration} can output the exact cardinality.
	
\end{myproof}

\noindent\textbf{\underline{Time Complexity Analysis}} 
The complexity of Post-ordered cardinality retrieval is upper bounded by $O(|\mathcal{V}|*|\mathcal{E}^Q|*|\mathcal{E}|)$ (in DAG scenario), where $|\mathcal{E}^Q|$ refers to the number of hyper-relational edges in query $Q$. The post-order iteration to search candidates takes $O(|\mathcal{E}^Q|*|\mathcal{E}|)$ time, where the iteration number is upper bounded by $O(|\mathcal{E}^Q|)$ and the search for each edge in $Q$ takes at most $O(|\mathcal{E}|)$. The search process for candidates at each leaf node $e_l$ takes at most $O(|\mathcal{V}|)$ time. In actual implementation, since we build indexes of each node's connected neighbors by the edge label in HKG, the actual running time is must less than $O(|\mathcal{V}|*|\mathcal{E}^Q|*|\mathcal{E}|)$. As for the cyclic queries contain at most 1 cycle, the extra checking condition can be executed in constant time.

\subsection{Generated Queryset Data Statistics}

As shown in Table~\ref{statistics}, we obtain the queryset on three wide-used HKGs, i.e., JF17K \cite{JF17K}, wikipeople \cite{wikipeople} and WD50K \cite{StarE}. These generated datasets on all three HKGs are diverse enough to cover all five query patterns with fact sizes 1,2,3,6,9,12. Also, our querysets cover the cardinality range from $<=10$ to $>10^5$, and consider a different number of bounded nodes.

\section{A Qualifier-aware GNN}\label{sec:model}
In this section, we introduce our GNN-based hyper-relational query encoder. We first introduce how to initialize the embedding for nodes and edges in the query and propose a qualifier-aware GNNs to learn a representation of the query to estimate the cardinality.
%Give a initialized HRQ, our encoder project the query into a latent space, which can be used to retrieve the cardinality value by an decoder module. Our encoder serves as an alternative to replace the encoder in existing learning-based CE frameworks to enhance the representational capacity for HRQs.  

\begin{figure*}[t]
	\centering
 
	\includegraphics[width=\linewidth]{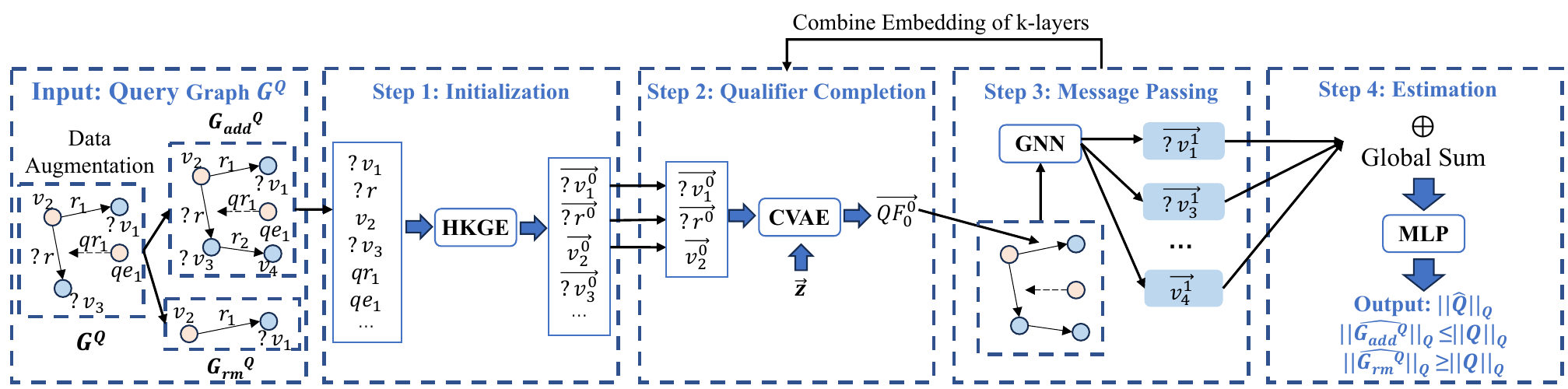}
	\vspace{-10pt}
	\caption{
		Given a query $Q$ with graph form $G^Q$, all atoms in $G^Q$ are firstly initialized to embedding. Then initialized $G^Q$ are fed into our qualifier-aware GNN encoder. Specifically, $G^Q$ is passed to $K$ message passing layers, In each layer, $G^Q$ will perform message passing after qualifier completion on each fact pattern. The node embeddings of $G^Q$ at different layer will be adaptively combined to generate the final representation. Finally, an MLP decoder computes the estimated cardinality based on query representation. In training phase, a data augmentation strategy modifies $G^Q$ edges/qualifiers first to generate augmented training data. We keep the relative magnitude between predicted cardinality of augmented query graph and cardinality of $Q$ cardinality of $Q$ in training. }
	\label{fig:framework}
	\vspace{-10pt}
\end{figure*}

\subsection{HKG Query Embedding Initialization} \label{sec:ini}
	We first initialize the embeddings for the nodes and edges in each query. 
	Specifically, for fixed nodes and edges in query that correspond to nodes and edges in HKG, we pretrain a StarE model following~\cite{StarE} to obtain nodes and edges embedding for each HKG.
	For variable nodes and edges, inspired by~\cite{GNCE}, we set the first dimension to be a unique numerical ID (e.g., $?v1$ to $ID1$, $?v2$ to $ID2$) and the remaining dimensions are set to 0.
	If the qualifier node/edge is a variable, we set all its remaining dimensions to 1, because we adopt rotate computation for each qualifier pair in Section 4.2. 
%	The rotation between a 1 vector and a non-zero vector remain the same for the non-zero one, which is in line with variable in nature and maintain the representation for non-variable atom in the qualifier~\cite{GNCE}. 
	After initialization, for any fact pattern $f^q = (s^q, p^q, o^q, \mathcal{Q}\mathcal{F}_f^q)$ in HRQ (we omit subscript $q$ in atoms embedding $h$ for simplicity), we obtain the representation $\mathbf{h}_{s}^{(0)}, \mathbf{h}_{p}^{(0)}, \mathbf{h}_{o}^{(0)}, \mathbf{h}_{qr}^{(0)}$ and $\mathbf{h}_{qe}^{(0)}$ ($(qr, qe) \in \mathcal{Q}\mathcal{F}_f^q$).

\begin{algorithm}[t]
	
	\KwIn{HKG $\mathcal{G}$ and generated query $Q$}
	\KwOut{Cardinality $\Vert Q \Vert$ of $Q$}
	
	$d \leftarrow \emptyset$ \\
	\uIf {$Q$ is a single branch without join} {
		find $V_{e_l} \in \mathcal{G}$ \\ \label{startnode}
		initialize $d$, set $d[v_{e_l}] \in V_{e_l}$ to 1 \\
		\For{$(r_i, e_i) \in Q$} {
			$V_{e_{i-1}} \leftarrow \emptyset$ \\
			
			\For{$v_{e_i} \in V_{e_i}$} {
				$V_{e_{i-1}}.add(v_{e_{i-1}})$ if $(v_{e_{i-1}}, r_i, v_{e_{i}}) \in \mathcal{G}$.\\ \label{nexthop}
				$d[v_{e_(i-1)}] = d[v_{e_(i-1)}] + d[v_{e_(i)}] $ \\
				remove $v \in d$ if $v$ does not update \\
			}
		}
		%STATE   $\widehat{R}=\widetilde{D}^{-\frac{1}{2}}(R+I)=\widetilde{D}^{-\frac{1}{2}}$ 
	}
	\uElse {
		\For {each branch $Q_i \in Q$ in post-order} {
			$d_i$ = CardinalityRetrieval($\mathcal{G}$, $q_i$, $d$) \\
			\For {$v \in d$} {
				\uIf {$v \in d_i$} {
					$d[v] = d[v] * d_i[v]$ \\
				}	
				\uElse {
					remove $v$ \\
				}
			}	
		}
	}
	$||q|| \leftarrow \sum_{v_{e_0} \in e_0} d[v_{e_0}]$ \\
	\textbf{return}  $\Vert Q \Vert$
	\caption{Cardinality Computation}
	\label{alg:querygeneration}
\end{algorithm}
\subsection{HKGs Query Encoder} \label{QualAgg}

\subsubsection{Qualifier-aware Message Passing}

As mentioned in Section~\ref{HKG}, qualifiers provide information about the main fact. Thus, we need to incorporate the information from qualifiers into the representations of each node and edge. 
We follow the design of StarE~\cite{StarE} to build our query encoder.
First, given a fact $f^q = (s^q, p^q, o^q, \mathcal{QF}_f^q)$, we learn the representation of qualifiers 

\begin{equation} \label{qual_aggregation}
	\mathbf{\tilde{h}}^{(k)}_{\mathcal{Q}\mathcal{F}_f} =\mathbf{ W}^{(k-1)}_{qual} \sum_{(qr, qe) \in \mathcal{Q}\mathcal{F}^q} \zeta(\mathbf{h}^{(k-1)}_{qr}, \mathbf{h}^{(k-1)}_{qe}),
\end{equation}
where $\mathbf{W}^{(k-1)}_{qual}$ is a trainable projection matrix shared by all qualifiers in query and $\zeta(\cdot)$ is a composition function between each qualifier pair embedding $\mathbf{h}_{qr}$ and $\mathbf{h}_{qe}$, where we use rotate operation following~\cite{StarE}. 
Note that since the qualifier information $\mathcal{QF}_f^q$ in each query fact $f^q$ may be incomplete, thus we propose to generate the incomplete qualifier information $\mathbf{\hat{h}}^{(k)}_{{QF}_{f,l}}=\mathcal{M}_\phi(f^q,k) $, where  $\mathcal{M}_\phi(\cdot)$ is a pretrained generative model and can infer the additional qualifier information given $f^q$. The details is in Section~\ref{ssec:qualifier_completion}.
Thus, the accurate qualifier embedding is as follows:
\begin{align}\label{eq:lambda}
	\mathbf{h}^{(k)}_{\mathcal{Q}\mathcal{F}_f^q}=	(1-\lambda) \cdot \mathbf{\tilde{h}}^{(k)}_{\mathcal{Q}\mathcal{F}_f^q} + \lambda\cdot\mathbf{\hat{h}}^{(k)}_{\mathcal{QF}_{f,l}^q}
\end{align}

Then, we use a $\gamma(\cdot)$ function~\cite{StarE} that combines the qualifiers $\mathbf{h}_{\mathcal{Q}\mathcal{F}_f^q}^{(k)}$ to the main relation $\mathbf{h}^{(k)}_{p}$, i.e., $\gamma(\mathbf{h}^{(k)}_r,\mathbf{h}^{(k)}_{\mathcal{Q}\mathcal{F}_f^q})= \mathbf{h}^{(k)}_r +  \mathbf{h}^{(k)}_{\mathcal{Q}\mathcal{F}_f^q}$.
%Also, we consider how to infer incomplete qualifiers of query to learn more accurate qualifier representation in Section~\ref{ssec:qualifier_completion}.
Then,
we adopt GIN layers~\cite{GIN} as our basis encoder and add the qualifier combination to learn the representation for each subject $s$ (same for object $o$) as follows:
\begin{equation} \label{qualifer attached message passing}
	\begin{split}
		\mathbf{h}_{s}^{(k)} = m_{\theta}^{(k-1)}(\mathbf{h}_{s}^{(k-1)} + \mathbf{h}_{s_{in}}^{(k-1)} + \mathbf{h}_{s_{out}}^{(k-1)})
	\end{split}
\end{equation}

\begin{equation} \label{income_message passing}
		\mathbf{h}_{s_{in}}^{(k)} =\sum_{o_{f_i} \in \mathcal{N}^+(s)}\sigma(\textbf{W}_e^{(k-1)}(\mathbf{h}_{s}^{(k-1)}||\gamma(\mathbf{h}_{r_{f_i}}^{(k-1)}, \mathbf{h}_{\mathcal{QF}_{f_i}}^{(k-1)})||\mathbf{h}_{o_{f_i}}^{(k-1)})) \nonumber
\end{equation}

\begin{equation} \label{outcome_message passing}
		\mathbf{h}_{s_{out}}^{(k)} =\sum_{o{f_g} \in \mathcal{N}^-(s)}\sigma(\textbf{W}_e^{(k-1)}(\mathbf{h}_{o_{f_g}}^{(k-1)}||\gamma(\mathbf{h}_{r_{f_g}}^{(k-1)}, \mathbf{h}_{\mathcal{QF}_{f_g}}^{(k-1)})||\mathbf{h}_{s}^{(k-1)})) \nonumber
\end{equation}

%In the message passing function, the major update is we replace the edge label embedding $\mathbf{h}_r$ to $\gamma(\mathbf{h}_r, \mathbf{h}_{\mathcal{Q}})$ that injects the qualifiers into the hyper-relational edge. 
%We borrow the idea from TPN message passing that splits the message from incoming edges and outgoing edges to target node $x_i$ and aggregates the main triple embedding by concatenation (||) via a dimensionality projection matrix $\textbf{W}$ to project the concatenated embedding to the same dimension as $\mathbf{h}_{s}$. 
where $m_{\theta}^{(k-1)}$ is an MLP that receives the updated message of node $s$ based on the setting of GIN layer and $\sigma$ is the $ReLU$ activation function. Also, we update the embedding of relation $r$ and items in each qualifier $(qr,qv) \in Q$ based on a transform matrix as $\mathbf{h}_r^{k}=\sigma(\mathbf{W}_r^{(k-1)}\mathbf{h}_r^{(k-1)})$, 
$\mathbf{h}_{qr}^{k}=\sigma(\mathbf{W}_{qr}^{(k-1)}\mathbf{h}_{qr}^{(k-1)})$, and $\mathbf{h}_{qe}^{k}=\sigma(\mathbf{W}_{qe}^{(k-1)}\mathbf{h}_{qe}^{(k-1)})$.
All mentioned matrices will be jointly trained according to Sec.\ref{sec:dataaugmentation}.

\subsubsection{Qualifier Completion}\label{ssec:qualifier_completion}

Recall that the qualifier information for a query fact in the HKGs is important to infer cardinality, which can provide context information for queries. For example, two queries $(?s, p, o_1)$ and $(?s, p, o_2)$ differ only in one entity $o_1$ and $o_2$, but their meanings can be significantly different based on qualifier information.
Therefore, we propose to pretrain a conditional variational autoencoder (CVAE) \cite{gnndis, lagnn} to generate the incomplete qualifier information $\mathbf{\hat{h}}^{(k)}_{\mathcal{QF}_{f,l}^q}$ at the $k$-th layer that does not appear in each query fact $f^q=(s^q,p^q,o^q,\mathcal{QF}_f^q)$. 

In general, the CVAE model consists of one encoder $p_{\phi_1}(\cdot)$ and one decoder $d_{\phi_2}(\cdot)$. 
Given input data $f=(s,p,o,\mathcal{QF}_{f,p})$ with partial qualifier information $\mathcal{QF}_{f,p}$ and the layer $k$, we first get the input vector as $\mathbf{x}= \mathbf{h}^{(k)}_s||\mathbf{h}^{(k)}_p||\mathbf{h}^{(k)}_o||\mathbf{\tilde{h}}^{(k)}_{\mathcal{QF}_{f,p}}$ where $\mathbf{\tilde{h}}^{(k)}_{\mathcal{QF}_{f,p}}$ is learned based on $\mathcal{QF}_{f,p}$ in Equation~\eqref{qual_aggregation}. 
Then the encoder $p_{\phi_1}(\cdot)$ will learn the latent representation for the input as $\boldsymbol{z}\sim p_{\phi_1}(\boldsymbol{z}|\textbf{x})$. 
Then, the decoder will generate the incomplete qualifier information $\mathbf{y} \sim p_{\phi_2}(\mathbf{\hat{h}}^{(k)}_{\mathcal{QF}_{f,l}}|\mathbf{x},\mathbf{z})$.

Specifically, we discuss two ways of constructing training input $\mathbf{x}$ and the output $\mathbf{y}$ based on existing facts in HKG $\mathcal{G}$ to pretrain the CVAE model. 
First, given each fact $f=(s,p,o,\mathcal{QF}_f)$ in a HKG, we first sample partial qualifiers $\mathcal{QF}_{f,p}$ and then use the information $f=(s,p,o,\mathcal{QF}_{f,p})$ to predict the incomplete qualifier information $\mathcal{QF}_{f,l}=\mathcal{QF}_f\setminus \mathcal{QF}_{f,p}$. 
In this case, under the $k$-th layer, we use the input $\mathbf{x}= \mathbf{h}^{(k)}_s||\mathbf{h}^{(k)}_p||\mathbf{h}^{(k)}_o||\mathbf{\tilde{h}}^{(k)}_{\mathcal{QF}_{f,p}}$ to predict $\mathbf{y}=\mathbf{\hat{h}}^{(k)}_{\mathcal{QF}_{f,l}}$, where $\mathbf{\tilde{h}}^{(k)}_{\mathcal{QF}_{f,p}}$ and $\mathbf{\hat{h}}^{(k)}_{\mathcal{QF}_{f,l}}$ are learned based on $\mathcal{QF}_{f,p}$ and $\mathcal{QF}_{f,l}$ in Equation~\eqref{qual_aggregation}. 
Second, we randomly replace some entity/relation in $f=(s,p,o,\mathcal{QF}_f)$ with variables, such as replacing $s$ with $?s$, and then use $(?s,p,o)$ to predict the information $\mathcal{QF}_f$. 
In this case, under the $k$-th layer, we use the input $\mathbf{x}= \mathbf{h}^{(k)}_s||\mathbf{h}^{(k)}_p||\mathbf{h}^{(k)}_o||\mathbf{0}$ to predict $\mathbf{y}=\mathbf{\hat{h}}^{(k)}_{\mathcal{QF}_{f}}$.
Finally, the evidence lower bound (ELBO)~\cite{gnndis, lagnn} is used for CVAE optimization.

%{\color{blue}
	\begin{theorem}
	Given a hyper-relational fact  $f = (s, p, o, \mathcal{QF}_{f,p})$,
	the input 
	 $\mathbf{x} = \mathbf{h}^{(k)}_s || \mathbf{h}^{(k)}_p || \mathbf{h}^{(k)}_o || \mathbf{\tilde{h}}^{(k)}_{\mathcal{QF}_{f,p}}$ represent the concatenation of the embedding
	$\mathbf{h}^{(k)}_s$, $\mathbf{h}^{(k)}_p$, and $\mathbf{h}^{(k)}_o$ of $s$, $p$, and $o$ at the $k$-th layer, respectively.
	Also, let
	$\mathbf{\tilde{h}}^{(k)}_{\mathcal{QF}_{f,p}}$ represents the embedding of the known qualifiers $\mathcal{QF}_{f,p}$ in the $k$-th GNN layer and
	let $\mathbf{y}= \mathbf{\hat{h}}^{(k)}_{\mathcal{QF}_{f,l}}$ denote the estimated embedding of the incomplete qualifiers $\mathcal{QF}_{f,l}$ at the $k$-th layer.
	Given a conditional variational encoder (CVAE)~\cite{gnndis, lagnn}  that consists of  
	an encoder $p_{\phi_1}(\cdot)$, a decoder $p_{\phi_2}(\cdot)$, and a variational parameters $q_\theta(\cdot)$, 
	the loss function $ \mathcal{L}_{cvae}(\textbf{y}, \textbf{x}, \phi_1, \phi_2, \theta)$ on each training instance can be obtained as follows.
	 	\begin{equation} \label{eq:loss}
			-KL(q_{\theta}(\boldsymbol{z}|\textbf{y}, \textbf{x})||p_{\phi_1}(\boldsymbol{z}|\textbf{y}, \textbf{x})) 
			+ \frac{1}{J} \sum_{j=1}^J{\rm log} p_{\phi_2}(\textbf{y}|\textbf{x},\boldsymbol{z}^j)
			%	\end{split}
		\vspace{-2pt}
	\end{equation}
	where $\boldsymbol{z}^{j} = g_{\theta}(\textbf{x}, \textbf{y}, \epsilon^{j}), \epsilon^{j} \sim \mathcal{N}(0, I)$ is sampled from a standard Gaussian distribution, $g_{\theta}(\cdot)$ is a reparameterization function~\cite{cvae} and $J$ denotes the number of training hyper-relational fact patterns. 
	
	\end{theorem}

\begin{myproof} \label{appx:cvaederivation}
We follow the general derivation for CVAE to derive ELBO~\cite{lagnn,gnndis}. For simplicity, we use $\textbf{x}$ to denote concatenated embedding of main triple and $\mathcal{Q}\mathcal{F}_{l}$. $\textbf{y}$ denoting prediction target $\mathcal{Q}\mathcal{F}_{p}$ while $J$ stands for the size of training data for CVAE and $\textbf{z}$ is a variable following normal distribution. In CVAE-based qualifier completer, the log likelihood ${\rm log} p_{\phi}(\textbf{y}|\textbf{x})$ should be maximized with lower bound as follows:
\begin{equation} \label{eq:likelihoodderive}
	\begin{split}
		&	{\rm log} p_{\phi}(\textbf{y}|\textbf{x}) 
		= \int q_{\theta}(\textbf{z}|\textbf{y}, \textbf{x}) ({\rm log}\frac{p_{\phi}(\textbf{y}, \textbf{z}|\textbf{x})}{q_{\theta}(\textbf{z}|\textbf{y}, \textbf{x})} +{\rm log}\frac{q_{\theta}(\textbf{z}|\textbf{y}, \textbf{x})}{p_{\phi}(\textbf{z}|\textbf{y}, \textbf{x})}){\rm d}\textbf{z}\\&
		= \int q_{\theta}(\textbf{z}|\textbf{y}, \textbf{x}){\rm log}\frac{p_{\phi}(\textbf{y}, \textbf{z}|\textbf{x})}{q_{\theta}(\textbf{z}|\textbf{y}, \textbf{x})}{\rm d}\textbf{z} + KL(q_{\theta}(\textbf{z}|\textbf{y}, \textbf{x})||p_{\phi}(\textbf{z}|\textbf{y}, \textbf{x}))\\&
		\geq  \int q_{\theta}(\textbf{z}|\textbf{y}, \textbf{x}){\rm log}\frac{p_{\phi}(\textbf{y}, \textbf{z}|\textbf{x})}{q_{\theta}(\textbf{z}|\textbf{y}, \textbf{x})}{\rm d}\textbf{z} \nonumber
	\end{split}
\end{equation}
By above Equation, we obtain ELBO of ${\rm log} p_{\phi}(\textbf{y}|\textbf{x})$. To maximize the log likelihood, we are supposed to maximize ELBO.

\begin{equation}
	\begin{split}
		&	ELBO
		= \int q_{\theta}(\textbf{z}|\textbf{y}, \textbf{x}){\rm log}\frac{p_{\phi}(\textbf{y}, \textbf{z}|\textbf{x})}{q_{\theta}(\textbf{z}|\textbf{y}, \textbf{x})}{\rm d}\textbf{z}\\&
		= \int q_{\theta}(\textbf{z}|\textbf{y}, \textbf{x}){\rm log}\frac{p_{\phi}(\textbf{z}|\textbf{x})}{q_{\theta}(\textbf{z}|\textbf{y}, \textbf{x})}{\rm d}\textbf{z} + \int q_{\theta}(\textbf{z}|\textbf{y}, \textbf{x}){\rm log}p_{\phi}(\textbf{y}|\textbf{x}, \textbf{z}){\rm d}\textbf{z}\\&
		= -KL(q_{\theta}(\textbf{z}|\textbf{y}, \textbf{x})||p_{\phi}(\textbf{z}|\textbf{y}, \textbf{x})) + \frac{1}{J}\sum_{j=1}^{J}{\rm log} p_{\phi}(\textbf{y}|\textbf{x},\textbf{z}^{(j)}) \nonumber
	\end{split}
\end{equation}

Thus, the loss function of CVAE is written as Equation~\eqref{eq:loss}.
%\vspace{-3pt}

\end{myproof}

After training, we can use the decoder of pre-trained CVAE to complete the missing qualifier pairs $\mathbf{\hat{h}}^{(k)}_{\mathcal{QF}_{f,l}^q}$. Given HRF pattern $f^q = (s^q, p^q, o^q, \mathcal{QF}^q_f)$ in HRQ, we can estimate the missing qualifier pairs $\mathbf{\hat{h}}^{(k)}_{\mathcal{QF}_{f,l}^q}$ through CVAE decoder by sampling a latent variable $\boldsymbol{z} \sim \mathcal{N}(0, I)$.

\subsubsection{Multi-layer Combination} \label{multilayer}
The number of layers of the GNN model affects the receptive field for each node. 
The fixed $2$-layered GNN in existing CE models restricts receptive fields, achieving unsatisfied performance.
A larger node's receptive field for nodes results in the representations for nodes being indistinguishable, which refers to an over-smoothing problem \cite{DEEPGNNref, DEEPGNN}. 
Thus,
we applied a trainable projection vector $\textbf{w} \in \mathbb{R}^{d_h \times 1}$  which is shared by all nodes to regularize the amount of information provided by each GIN layer to mitigate the over-smoothing problem. 
In detail, for each node representation $\mathbf{h}_e^{(k)} \in  \mathbb{R}^{d_h \times 1}$ in k-th GIN layer, the model obtains a retainment score by $\sigma(\textbf{w} \odot \mathbf{h}_e^{(k)})$ as an adaptive weight to control the amount of information used in the k-th layer.  
The final node representation of each node $e \in \{s,o\}$ in each fact pattern is computed as:
\begin{equation} \label{eq:gnnlayer}
	\vspace{-5pt}
	\mathbf{h}_e^{f} = \sum_{k=1}^{L} (\sigma(\textbf{w} \odot \mathbf{h}_e^{(k)}) \odot \mathbf{h}_e^{(k)}), \forall e \in \{s, o\}
\end{equation}
%By the above formula, the receptive field for nodes can be adaptively enlarged and differentiated across different nodes in the query. 
Then, we use the  global sum readout to summarize the final latent representation, which will be transformed by an MLP decoder to predict the cardinality of the query. 
The parameters of our model will be optimized by Mean Squared Error (MSE) between predicted cardinality and the logarithm of ground truth cardinality~\cite{GNCE}, i.e., $({\Vert Q \Vert}_{\mathcal{Q}} - {\Vert \hat{Q} \Vert}_{\mathcal{Q}})^2$.
%with logarithm omitted for simplicity. 

\noindent\textbf{\underline{Time Complexity of HRQE in cardinality estimation.}}
In this subsection, we mainly discuss the time complexity for HRQE over CE task. Suppose the number of nodes in hyper-relational query is $N$, the dimension of embedding is $D$, the average degree of nodes is $d_n$ and the average number of qualifier pairs is $d_q$. In general, while the time complexity of sampling based method takes quadratic time $O(N^2)$ to search the index of mapping between query nodes and corresponding KG nodes. In each layer, it takes $O(N*d_n)$ to aggregate the embedding of neighbor nodes and edges and takes $O(N*D^2)$ to update the embedding in GIN. The embeddings of qualifiers are aggregated to the embedding of corresponding relation for $d_q$ times in each edge. Thus, the time complexity is $O(N*d_n* dq + N*D^2)$ in each layer. Suppose the number of layers of HRQE is $L$, the total time complexity for HRQE on a single query is $O(L(N*d_n* d_q + N*D^2))$, which is linear in terms of $L$ and $N$. Thus, HRQE is less potential to have scalability problem as the number of GNN layers increasing or the number of query nodes growth.

\begin{table}[t]
	\centering
	\small
	%\vspace{-10pt}
	\caption{Statistics for three HKGs. }
	%\vspace{-10pt}
%	\setlength\tabcolsep{0.5pt}
%	\resizebox{\linewidth}{!}{
%		\renewcommand{\arraystretch}{2}
		\begin{tabular}{c|l|c|c|c}
			\hline
			\textbf{HKG}        & \textbf{\#Facts} & \textbf{Qual Facts(\%)}  
			& \textbf{\#Entities} & \textbf{\#Relations} \\  \hline
			\textbf{WD50K}      & 236507  & 13.6\%                                                     & 47156      & 532         \\  
			
			\textbf{Wikipeople} & 369866  & 2.6\%                                                    & 34839      & 375         \\ 
			
			\textbf{JF17K}      & 100947  & 45.9\%                                                   & 28645      & 322         \\ \hline
		\end{tabular}

%	}
	\label{tab:stat}
 
\end{table}

%\begin{table}[t]
%	\centering
%	\small
%	%\vspace{-10pt}
%	\caption{Statistics for three HKGs. }
%	%\vspace{-10pt}
%	\setlength\tabcolsep{0.5pt}
%	%	\resizebox{\linewidth}{!}{
%		%		\renewcommand{\arraystretch}{2}
%		\begin{tabular}{c|l|c|c|c|c}
%			\hline
%			\textbf{HKG}        & \textbf{Facts\#} & \textbf{Qual Facts(\%)} & \multicolumn{1}{l|}{\textbf{Multi Qual Facts\#}} & \textbf{Entities\#} & \textbf{Relations\#} \\ \hline
%			\textbf{WD50K}      & 236507  & 13.6\%         & 1600                                                        & 47156      & 532         \\ \hline
%			\textbf{Wikipeople} & 369866  & 2.6\%          & 130                                                         & 34839      & 375         \\ \hline
%			\textbf{JF17K}      & 100947  & 45.9\%         & 2267                                                        & 28645      & 322         \\ \hline
%		\end{tabular}
%		
%		
%		%	}
%	\label{tab:stat}
%	\vspace{-15pt}
%\end{table}

\subsection{Data Augmentation-based Model Training} \label{sec:dataaugmentation}

%Inspired by the success of GNCE~\cite{GNCE} over triple-KG, we apply the Multi Layer Perceptron (MLP) as the decoder to estimate the cardinality from encoded query embedding, where encoder and decoder are joinly trained in supervised manner via Mean Squared Error (MSE) loss function.
%We compute the difference between the model output and the logarithm of the true cardinality.
% in order to stabilize training on the large range of cardinality.

Besides MSE loss, we also designed a simple yet effective data augmentation strategy due to the scarcity of training data and enhance the model generalizability.
For each query $Q$ in training set, we build two auxiliary query sets, namely $Q_{\textit{add}}$ and $Q_{\textit{rm}}$. 
The query in $Q_{\textit{add}}$ is obtained by adding an edge/a qualifier to $Q$.
Intuitively, adding an extra edge or qualifier to \( Q \) means the cardinality of the query in \( Q_{\textit{add}} \) should be less than or equal to that of \( Q \).
Inversely, the query in $Q_{\textit{rm}}$ is obtained by removing an edge/a qualifier to $Q$, whose cardinality should be greater than or equal to the cardinality of $Q$, since the removing operation increase the matched subgraphs. 

Thus, we can develop an augmented loss function $ \mathcal{L}_{CE}$ ({${\Vert \hat{Q} \Vert}_{\mathcal{Q}}$ and ${\Vert \hat{Q}^{\prime} \Vert}_{\mathcal{Q}}$ stand for model output) based on the cardinality magnitude relationship as following:
\begin{equation}
	\begin{split}
			({\Vert Q \Vert}_{\mathcal{Q}} - {\Vert \hat{Q} \Vert}_{\mathcal{Q}})^2 & 
			+ \frac{1}{\vert Q_{\textit{add}} \vert} \sum_{Q^{\prime} \in Q_{\textit{add}}} ReLU({\Vert{\hat{Q}^{\prime}}\Vert}_{\mathcal{Q}}-{\Vert Q\Vert}_{\mathcal{Q}})\\& 
			+ \frac{1}{\vert Q_{\textit{rm}} \vert} \sum_{Q^{\prime} \in Q_{\textit{rm}}} ReLU({\Vert Q \Vert}_{\mathcal{Q}}-{\Vert{\hat{Q}^{\prime}}\Vert}_{\mathcal{Q}}) \\&  \nonumber
	\end{split}
\end{equation}
In data augmented training, we only stress the relative magnitude between predicted cardinality in auxiliary query sets and the true cardinality of given instance. Thus, we add a $ReLU$ function to avoid the model falsely outputs the same value for auxiliary query sets and the given instance.

\begin{table*}[t]
	\centering
	\small
	
	\caption{Effectiveness (Mean q-Errors) and efficiency (time with seconds) over three HKGs. '-' indicates the mean q-error is higher than $10^{20}$ thus we do not show the exact value.
		The \textbf{bold number} and the \underline{underline number} indicate the best and the second best performance, respectively.	
	}
	\vspace{-5pt}
 	\setlength\tabcolsep{1pt}
	\label{tab:mainexp}
	\resizebox{\linewidth}{!}{
			\renewcommand{\arraystretch}{1.3}
	\begin{tabular}{c|cccc|cccc|cccc}
		\hline
		  \multirow{2}{*}{\textbf{Model}} & \multicolumn{4}{c|}{\textbf{Mean q-Errors}}                                   & \multicolumn{4}{c|}{\textbf{Inference Time (s)}}     & \multicolumn{4}{c}{\textbf{Training Time (s)}}                                  \\ \cline{2-13} 
		                    &  \textbf{WD50K} & \textbf{Wikipeople} &\textbf{JF17K} & \textbf{WD50K-QE}   &  \textbf{WD50K} & \textbf{Wikipeople} &\textbf{JF17K} & \textbf{WD50K-QE}  &  \textbf{WD50K} & \textbf{Wikipeople} &\textbf{JF17K} & \textbf{WD50K-QE}  \\ \hline
		
		  IMPR~\cite{IMPR}                      & \multicolumn{1}{c}{$1.35\cdot10^{7}$}      & \multicolumn{1}{c}{ $8.69\cdot10^{6}$}           &$7.25\cdot10^{5}$ & $6.25\cdot10^{5}$      & \multicolumn{1}{c}{$1.35\cdot10^{-2}$}      & \multicolumn{1}{c}{$1.3\cdot10^{-2}$}           &  $1.34\cdot10^{-2}$ &  $9.12\cdot10^{-3}$  &      \\  
		  JSUB~\cite{JSUB}                      & \multicolumn{1}{c}{$1.34\cdot10^7$}      & \multicolumn{1}{c}{$3.10\cdot10^{18}$}           & - & $3.29\cdot10^5$      & \multicolumn{1}{c}{$1.5\cdot10^{-2}$}      & \multicolumn{1}{c}{$1.12\cdot10^{-2}$}           & - &  $9.76\cdot10^{-3}$    & \multicolumn{4}{c}{No training required}    \\ 
		  WanderJoin~\cite{wanderjoin}                   & \multicolumn{1}{c}{$1.31\cdot10^7$}      & \multicolumn{1}{c}{-}           & -  &$2.91\cdot10^5$     & \multicolumn{1}{c}{$1.12\cdot10^{-2}$}      & \multicolumn{1}{c}{-}           & -&  $8.33\cdot10^{-3}$    &      \\ 
		  ALLEY~\cite{ALLEY}                   & \multicolumn{1}{c}{$9.89\cdot10^6$}      & \multicolumn{1}{c}{$7.47\cdot10^{10}$}           & -   & $1.45\cdot10^5$    & \multicolumn{1}{c}{$1.36\cdot10^{-2}$}      & \multicolumn{1}{c}{$1.56\cdot10^{-2}$}           & - &  $1.03\cdot10^{-2}$    &      \\ \hline

		 QTO~\cite{qto}                 & \multicolumn{1}{c}{-}      & \multicolumn{1}{c}{-}           &   -  &   -   & \multicolumn{1}{c}{\textbf{-}}      & \multicolumn{1}{c}{\textbf{-}}           &  \textbf{-} &   - & \multicolumn{1}{c}{\textbf{-}} & \multicolumn{1}{c}{\textbf{-}} & \textbf{-}  &   - \\  \cline{1-13}
		%		&  UnRavL~\cite{unravl}                 & \multicolumn{1}{c|}{233943}      & \multicolumn{1}{c|}{120234}           &   702934     & \multicolumn{1}{c|}{\textbf{13}}      & \multicolumn{1}{c|}{\textbf{11}}           &  \textbf{11}  & \multicolumn{1}{c|}{\textbf{200}} & \multicolumn{1}{c|}{\textbf{122}} & \textbf{177}   \\ \cline{2-11}
		 StarQE~\cite{StarQE}                 & \multicolumn{1}{c}{$3.01\cdot10^{5}$}      & \multicolumn{1}{c}{$7.47\cdot10^{4}$}           &   $4.45\cdot10^{5}$ &   $2.58\cdot10^{4}$    & \multicolumn{1}{c}{\underline{$1.08\cdot10^{-3}$}}      & \multicolumn{1}{c}{\underline{$1.00\cdot10^{-3}$}}           &  \underline{$1.00\cdot10^{-3}$} &   \underline{$9.53\cdot10^{-4}$} & \multicolumn{1}{c}{\underline{$2.00\cdot10^{2}$}} & \multicolumn{1}{c}{\underline{$1.22\cdot10^{2}$}} & \underline{$1.77\cdot10^{2}$} &   \underline{$1.15\cdot10^{2}$}  \\ 
		 StarQE~\cite{StarQE}+GIN~\cite{GIN}               & \multicolumn{1}{c}{$8.72\cdot10^{3}$}      & \multicolumn{1}{c}{$1.70\cdot10^{5}$}           &   $1.32\cdot10^{3}$  &   \underline{$3.66\cdot10^{2}$}   & \multicolumn{1}{c}{$\textbf{1.08}\cdot\textbf{10}^{\textbf{-3}}$}      & \multicolumn{1}{c}{$\textbf{1.00}\cdot\textbf{10}^{\textbf{-3}}$}   &  $\textbf{1.00}\cdot\textbf{10}^{\textbf{-3}}$ &  $\textbf{9.81}\cdot\textbf{10}^{\textbf{-4}}$   & \multicolumn{1}{c}{$\textbf{1.82}\cdot\textbf{10}^{\textbf{2}}$} & \multicolumn{1}{c}{$\textbf{1.18}\cdot\textbf{10}^{\textbf{2}}$} & $\textbf{1.45}\cdot\textbf{10}^{\textbf{2}}$  &  $\textbf{1.03}\cdot\textbf{10}^{\textbf{2}}$   \\   \cline{1-13}
		 GNCE~\cite{GNCE}                          & \multicolumn{1}{c}{\underline{$1.66\cdot10^{4}$}}      & \multicolumn{1}{c}{\underline{$1.66\cdot10^{3}$}}          &   \underline{$5.74\cdot10^{2}$}  &   $6.33\cdot10^{3}$    & \multicolumn{1}{c}{$7.26\cdot10^{-2}$}      & \multicolumn{1}{c}{$7.17\cdot10^{-3}$}           & $7.33\cdot10^{-3}$ & $2.96\cdot10^{-3}$ & \multicolumn{1}{c}{$5.82\cdot10^{2}$} & \multicolumn{1}{c}{$1.77\cdot10^{2}$} & $1.77\cdot10^{2}$ &  $1.46\cdot10^2$     \\
		 GNCE~\cite{GNCE}+Qual                          & \multicolumn{1}{c}{$4.84\cdot10^{3}$}      & \multicolumn{1}{c}{$2.93\cdot10^{4}$}          &   $4.50\cdot10^{3}$ &   $6.38\cdot10^{3}$    & \multicolumn{1}{c}{$1.23\cdot10^{-2}$}      & \multicolumn{1}{c}{$1.02\cdot10^{-2}$}           & $1.09\cdot10^{-2}$  & $7.18\cdot10^{-3}$ & \multicolumn{1}{c}{$2.02\cdot10^{2}$} & \multicolumn{1}{c}{$1.87\cdot10^{2}$} & $1.89\cdot10^{2}$ &  $1.77\cdot10^{2}$     \\ \cline{1-13} 
		 \textbf{HRQE (ours)}                    & \multicolumn{1}{c}{$\textbf{2.97}\cdot\textbf{10}^{\textbf{2}}$}      & \multicolumn{1}{c}{$\textbf{1.35}\cdot\textbf{10}^{\textbf{3}}$}           & $\textbf{3.55}\cdot\textbf{10}^{\textbf{2}}$ & $\textbf{2.70}\cdot\textbf{10}^{\textbf{1}}$     & \multicolumn{1}{c}{$5.83\cdot10^{-3}$}      & \multicolumn{1}{c}{$4.92\cdot10^{-3}$}           &  $5.50\cdot10^{-3}$ &  $1.85\cdot10^{-3}$  & \multicolumn{1}{c}{$4.93\cdot10^{2}$} &  \multicolumn{1}{c}{$3.40\cdot10^{2}$}   & $4.28\cdot10^{2}$ & $2.86\cdot10^{2}$   \\ \hline
	\end{tabular}
	
}
\end{table*}
%\vspace{-15pt}

\section{Experiments}\label{sec:experiments}

In this section, we compare  our proposed HRQE  with the state-of-the-art baselines to address the following research questions:
%Section~\ref{ssec:exp_setting} details the experimental settings. Section~\ref{ssec:main_results} presents the results regarding effectiveness and efficiency. Additionally, Section~\ref{ssec:ablation} covers ablation studies, Section~\ref{ssec:para} includes parameter sensitivity and Section~\ref{exp:inductive} discusses inductives evaluations .
\begin{itemize}[leftmargin=*]
	\item \textbf{RQ1:}How does HRQE compare to other state-of-the-art baselines in terms of both effectiveness and efficiency?

	%providing a comprehensive benchmark for cardinality estimation on HKGs.
	
	\item \textbf{RQ2:} How do the components (i.e., qualifier-aware message passing, data augmentation strategy and qualifier aggregation function) in HRQE affect the overall performance? 
	
	\item \textbf{RQ3:} How do hyper-parameters $L$ in Equation~\eqref{eq:gnnlayer} and $\lambda$ in Equation~\eqref{eq:lambda} affect the effectiveness of HRQE?
	
	%{\color{blue}\item We propose a simple yet effective data augmentation strategy to keep cardinality estimation consistency over the whole queryset by the relative cardinalities between queries.}
	
	\item \textbf{RQ4:} How effectively does HRQE perform under different query patterns, fact sizes and qualifier with incomplete qualifiers?
	
	%on the newly constructed diverse and unbiased HKG datasets.
	\item \textbf{RQ5:} How effectively does HRQE generalize to queries containing elements not encountered during the training phase?
 
\end{itemize}

\subsection{Experiment Setting}\label{ssec:exp_setting}

\subsubsection{Datasets and Metrics} We conduct experiments over three popular HKGs: JF17K \cite{JF17K}, wikipeople \cite{wikipeople} and WD50K \cite{StarE}. Table~\ref{tab:stat} summarizes the statistics about the HKGs. 
The statistics for generated querysets have been provided in Table~\ref{statistics}. 
The training and testing data are split with ratio 6:4.
Following~\cite{GNCE,LSS,GCARE},
	 we evaluate the effectiveness of CE approaches with respect to q-error $q-error(Q) = max(\frac{||Q||_{\mathcal{Q}}}{{||\hat{Q}||_{\mathcal{Q}}}}, \frac{{||\hat{Q}||_{\mathcal{Q}}}}{||Q||_{\mathcal{Q}}})$,
    where  $||Q||_{\mathcal{Q}}$ and $||\hat{Q}||_{\mathcal{Q}}$ are the ground cardinality and predicted cardinality, respectively. 
	We report the average q-error for test queries in the main experiment. %We also draw the boxplots to compare the q-error with outliers removed by matplotlib between different CE methods.

All experiments are conducted on CentOS 7 with a 20-core Intel(R) Xeon(R) Silver4210 CPU@2.20GHz, 8 NVIDIA GeForce RTX 2080 Ti GPUs (11G), and 92G of RAM.
%where the codes are available in https://anonymous.4open.science/r/HRQE-D2E5/

\subsubsection{Baselines} 
%Since no existing cardinality estimation baselines over HKGs, we simply run existing methods on querysets with qualifiers ignored. 
We compare and comprehensively investigate our method to 9 representative baselines from sampling-based approaches and learning-based methods. 
%%Except for existing CE works, since there are several hyper-relational query embedding models~\cite{StarQE, NQE, galkin2022inductive, efocqa} proposed for hyper-relational question-answering problem which also encodes hyper-relational query into a vector, they could be alternative baselines to compare with our work.i
%In recent years, there are multiple query encoders proposed for question-answering task that project query into an embedding and retrieve answer entities from the embedding. For triple-formed query encoder, we include QTO~\cite{qto} in our main experiment.
%We directly utilize the output of cardinality prediction head of QTO to predict queries' cardinality. 
%As for hyper-relational query encoder, we select StarQE~\cite{StarQE} which built upon CompGCN~\cite{compgcn} layers with qualifier aggregation module. We combine StarQE encoder with our MLP decoder and train on three querysets separately.
%To further evaluate the performance gain for HRQE, we simply modified StarQE~\cite{StarQE} by replacing CompGCN encoder to GIN encoder while maintaining the qualifier aggregation module in the encoder. We also modified GNCE~\cite{GNCE} by adding a qualifier aggregation module in its GIN message passing layers, namely StarQE+GIN and GNCE+Qual respectively.
%%We do not contains the state-of-the-art sampling-based method \textit{alley} \cite{ALLEY} due to the implementation failed over HKG. 

%{\color{blue}
\begin{itemize}[leftmargin=*]
	\item \textbf{Sampling-based methods.} They sample corresponding sub-graphs from KG and estimate the cardinality by probability of each sampled sub-graph. We compare our model with 4 methods, {\textit{IMPR}}~\cite{IMPR}, \textit{JSUB}~\cite{JSUB}, \textit{WanderJoin}~\cite{wanderjoin}, and \textit{ALLEY}~\cite{ALLEY}.  
	%providing a comprehensive benchmark for cardinality estimation on HKGs.
	
	\item \textbf{QTO~\cite{qto}.}
	 It is a KG query encoder to find the answer entity by tree optimization which can predict the cardinality of query. 
	
	\item \textbf{StarQE~\cite{StarQE}.} 
	It is a HKG query encoder built upon CompGCN~\cite{compgcn} with qualifier aggregation module. We combine StarQE encoder with MLP decoder to output the cardinality of each query. 
	We set a variant called \textit{StarQE~\cite{StarQE}+GIN~\cite{GIN}}
	by replacing CompGCN~\cite{compgcn} to GIN~\cite{GIN}. 
	We compare both \textit{StarQE} and \textit{StarQE+GIN }.
	
	%{\color{blue}\item We propose a simple yet effective data augmentation strategy to keep cardinality estimation consistency over the whole queryset by the relative cardinalities between queries.}
	
	\item \textbf{GNCE~\cite{GNCE}.} It is the-state-of-the-art cardinality estimator of KG queries that employs RDF2Vec~\cite{RDF2Vec} initialization embedding and GIN layers. We also add the qualifier aggregation function into its message passing scheme, called \textit{GNCE+Qual}. 
	We compare both \textit{GNCE} and \textit{GNCE+Qual }with our model.
	
	\item \textbf{HRQE (Ours).}
	 We propose a novel qualifier-aware  GNN model that incorporates qualifier information, 
	 adaptively combines outputs from multiple GNN layers, and utilizes data augmentation for accurate cardinality estimation on HKGs.

\end{itemize}

\subsubsection{Hyperparameter Setting} 
For sampling baselines, we maintain all default settings from G-CARE~\cite{GCARE} and average the results over 30 runs. 
%For sampling-based methods, we follow the default setting f
For all learning-based baselines, we train one model per HKG and maintain all default settings with two GNN layers,
For our model, we set qualifier aggregation function $\zeta(\cdot)$ is set to rotate in default. 
We tune the GNN layer number $L$ in Equation~\eqref{eq:gnnlayer} as $L \in \{2, 3, 4, 5, 6, 7\}$. 
Then, we tune the qualifier completion weight $\lambda$ in Equation~\eqref{eq:lambda} as  $\lambda \in  \{0,0,1,0.2,\cdots 1\}$. 
We tune the two hyperparameters $L$ and $\lambda$ by grid search over three datasets separately. 
For the  variants of our model, the number of message passing layers is set to 5. 
The batch-size is 32 and epoch is 100 for all learning-based baselines and our HRQE model  with Adam optimizer \cite{ADAM}.

\subsection{Main Results} \label{ssec:main_results}
 %As for ablation study, hyperparameter sensitivity and inductive cases, we utilize boxplot for q-error among each group of queries to evaluate the effectiveness.
% \begin{figure}[h]
% 	\centering 	
% 	%\vspace{-10pt}
% 	\subfloat[WD50K]	{\centering\includegraphics[width = 0.32\linewidth]{figs/overall/wd50k_overall2_shape_boxplot.pdf}}
%% 	\subfloat[wikipeople]	{\centering\includegraphics[width = 0.32\linewidth]{figs/overall/wikipeople_overall2_shape_boxplot.pdf}}
% 	
% 	\hfill
% 	\subfloat[JF17K]	{\centering\includegraphics[width = 0.32\linewidth]{figs/overall/jf17k_overall2_shape_boxplot.pdf}}
% 	%	\label{ssec:tendency:tendecy}
% 	
% 	%\vspace{-10pt}
% 	%	\label{ssec:tendency:tendecy}
% 	\caption{Main experiment results grouping by query patterns over three WD50K and JF17K}
% 	%\vspace{-10pt}
% 	\label{fig:topoexp}
% 	
% \end{figure}
 
\subsubsection{Effectiveness Evaluation} 
 	We compare HRQE's effectiveness with baselines via average q-error over three constructed query sets and the existing WD50K-QE.
% 	The test queries are grouped by different query patterns and fact sizes in each query to demonstrate the generalizability across different type of queries. 
% 	This criteria can be used to evaluate the performance of qualifier completer, thus applied in Section.~\ref{ssec:qualcomple} as well.
% 	We provided the boxplot grouping different query patterns and incomplete qualifiers as fig~\ref{fig:topoexp} and fig~\ref{fig:incompleteexp}. We put figure~\ref{fig:factsize} fact size boxplot in appendix due to space limit.
	As compared in Table~\ref{statistics}, WD50K-QE is a much simpler dataset with limited cardinality range.  
    Thus, all methods have less q-error compared to their performance on our proposed query benchmarks. 
	As shown in Table \ref{tab:mainexp}, all sampling-based baselines achieve unpromising result on hyper-relational queries since the qualifiers in queries are not considered in sampling-based methods. The information loss causes increased possibility of sampling failure and estimation error.
	
	With respect to learning-based methods, the query encoder for triple KG, QTO~\cite{qto}, have effectiveness problem due to the discard of qualifiers. Besides, as the cardinality prediction head for QTO merely predicts the number of answer nodes which is actually inconsistent with the cardinality definition to hyper-relational queries, QTO has severe estimation error.
	
	As for StarQE, it is less effective because the backbone CompGCN is not suitable for subgraph-matching related tasks though it has achieved better performance on link prediction task in recent studies. Compared to StarQE, its variant StarQE+GIN achieved better performance as GIN is a more powerful model for graph counting tasks. The experiment results also in line with the findings of previous CE works~\cite{LSS,GNCE,neurSC} that GIN is the well-suited for CE task. 
	Compared to our model, HRQE outperforms StarQE+GIN significantly over four datasets. As HRQE and StarQE+GIN has same backbone GNN and qualifier-aware message passing, it illustrates that our proposed components, including CVAE qualifier completer and data augmentation strategy are effective components as well.

	 Compared to GNCE and its variant GNCE+Qual, our model HRQE outperforms them as well. Besides above listed components, the severe information loss for initial embedding RDF2Vec~\cite{RDF2Vec} of GNCE is also a reason of high q-error, showing that RDF2Vec~\cite{RDF2Vec} is not a suitable embedding in HKGs.
	
%	According to Table~\ref{tab:stat}, over 45.9\% of facts in JF17K contain qualifiers, but the number of qualifier pairs within one fact is fewer than in WD50K. This indicates there are relatively fewer chances for queries in JF17K to have incomplete qualifiers. Figure~\ref{fig:lambda} shows that $\lambda=0.5$ is the performance peak, consistent with qualifier statistics over WD50K and JF17K. 
% Besides, the multi message passing layers enlarge the receptive fields of nodes and improve the capability for complex query patterns, such as petal and flower queries showing in Figure~\ref{fig:topoexp}. 
	{\color{blue}
%\begin{table}[h] 
%	\centering
%	
%	
%	
%	\caption{Effectiveness (Mean q-Errors) for learning-based models over WD50K-QE. 
%		The \textbf{bold number} and the \underline{underline number} indicate the best and the second best performance, respectively.	
%	}
%	\vspace{-10pt}
%	\label{tab:wd50kqe}
%	\begin{tabular}{c|c}
%		\hline
%		\multirow{1}{*}{\textbf{Model}} & \multicolumn{1}{c}{\textbf{Mean q-Errors}}                          \\ \cline{1-2} 
%		
%		StarQE+GIN                      & \multicolumn{1}{c}{\underline{$3.66\cdot10^{2}$}}      \\ \hline
%		GNCE                      & \multicolumn{1}{c}{$6.33\cdot10^{3}$}        \\ \hline
%		GNCE+Qual                      & \multicolumn{1}{c}{$6.38\cdot10^{3}$}          \\ \hline
%		\textbf{HRQE (ours)}                    & \multicolumn{1}{c}{$\textbf{2.70}\cdot\textbf{10}^{\textbf{1}}$}         \\ \hline
%	\end{tabular}
%	\vspace{-10pt}
%\end{table}
	
%	Furthermore, we also compare the learning-based models on existing query dataset, WD50K-QE~\cite{StarQE}. 
%	
%	HRQE maintains its leading performance about estimation accuracy, illustrating that our model can adapt simple queries well. 
%	Furthermore, the proposed data augmentation strategy also helps the model to learn the structure of different queries.
%
}
\subsubsection{Efficiency Evaluation} 
%\footnote{\# add table number}
{\color{black}In terms of efficiency, we report the average inference time for each query and training time for one epoch on three fore-mentioned querysets in Table~\ref{tab:mainexp}. 
Sampling-based approaches, such as ALLEY and WanderJoin, need abundant inference time since they need to sample answers compute the cardinality for each query.
Our model HRQE outperforms all sampling-based baseline methods, GNCE and its variant in terms of inference time. Though StarQE and its GIN variant is more time efficient since the simpler model structure, it suffers from severe effectiveness problem. Besides, time cost of HRQE is still comparable to StarQE and StarQE+GIN model in terms of average inference time per query.
 As for training, data augmentation strategy unavoidably roughly double the training time as more queries are computed gradients compared to learning baselines. However, the learning-based CE models are trained offline so that the increased training cost is still acceptable. HRQE costs more training time than GNCE+Qual and StarQE+GIN due to data augmentation strategy and increased number of GNN layers. 

}

\subsection{Ablation Study}\label{ssec:ablation}

{\color{black}In this section, we mainly study the effectiveness of each proposed components, including qualifier completer, data augmentation strategy and qualifier aggregation function, summarized in Table~\ref{tab:ablation}. }

%We study the impact of different components in our model, summarizing in Table~\ref{tab:ablation}.

\subsubsection{Qualifier Completer:} To illustrate the importance of qualifier completion, we set a variant HRQE\_NoQual which removes qualifier completer in message passing. As shown in Table~\ref{tab:ablation}, HRQE outperforms HRQE\_NoQual across all of three querysets for each type of query patterns, indicating that qualifier completion is essential to the query's cardinality, thus it is of importance to utilize qualifiers for accurately estimating the true cardinality of hyper-relational queries.

\subsubsection{Data augmentation:} 
The data augmentation technique in Section~\ref{sec:dataaugmentation} is to  alleviate data scarcity and increase model generalization.
We set HRQE\_NoAug that removes data augmentation in training. Figure~\ref{fig:topoexp}, Figure~\ref{fig:incompleteexp} and Figure~\ref{fig:factsize} illustrate that HRQE outperforms HRQE\_NoAug across three querysets, indicating that the proposed data augmentation strategy can indeed improve the model effectiveness in training phase.

\subsubsection{Qualifier aggregation function:} Besides, we also provide replace the qualifier aggregation function $\gamma$  in qualifier-aware message passing by concatenation and multiplication,  to study how qualifier aggregation function influences the estimation effectiveness, namely HRQE\_Concat and HRQE\_Multiply. HRQE beats HRQE-concat and HRQE-multiply underlying that summation is a more suitable choice of $\gamma$. 

Furthermore, to evaluate whether it works for which treat qualifiers as nodes attributions, we add a qualifier-decomposing variant, HRQE\_Decompose.  Specifically, given $f^q=(s^q,p^q,o^q,\{{qr}_j^q,{qe}_j^q\}_{j=1}^2)$, $f^q$ will be decomposed into $(s^q,p^q,o^q)\cap(s^q,{qr}_1^q,{qe}_1^q)\cap(s^q,{qr}_2^q,{qe}_2^q)$. However, HRQE outperforms the variant significantly over three datasets. It indicates that there will be unavoidable information loss if regarding qualifiers as nodes’ attributions, leading to the necessity of utilizing qualifier information by aggregation and completion. 
\begin{table}[t] 
	\centering

	\caption{Ablation study of Effectiveness (Mean q-Errors) for HRQE over three HKGs. 
		The \textbf{bold number} and the \underline{underline number} indicate the best and the second best performance, respectively.	
	}
	\vspace{-10pt}
	\label{tab:ablation}
	\begin{tabular}{c|ccc}
		\hline
		\multirow{2}{*}{\textbf{Model}} & \multicolumn{3}{c}{\textbf{Mean q-Errors}}                          \\ \cline{2-4} 
		& \multicolumn{1}{c}{\textbf{WD50K}} & \multicolumn{1}{c}{\textbf{Wikipeople}} & \textbf{JF17K}  \\ \hline
		
		\textbf{HRQE\_NoQual}                    & \multicolumn{1}{c}{$1.36\cdot10^{3}$}      & \multicolumn{1}{c}{$1.91\cdot10^{4}$}           &   \underline{$5.70\cdot10^{2}$}     \\ 
		\textbf{HRQE\_NoAug}                      & \multicolumn{1}{c}{\underline{$3.42\cdot10^{2}$}}      & \multicolumn{1}{c}{\underline{$1.86\cdot10^{3}$}}           &  $6.80\cdot10^{2}$      \\ 
		\textbf{HRQE\_Concat}                      & \multicolumn{1}{c}{$2.12\cdot10^{3}$}      & \multicolumn{1}{c}{$3.07\cdot10^{3}$}           &  $5.59\cdot10^{2}$      \\ 
	\textbf{HRQE\_Multiply}                      & \multicolumn{1}{c}{$3.14\cdot10^{3}$}      & \multicolumn{1}{c}{$8.93\cdot10^{5}$}           &  $2.13\cdot10^{3}$      \\ 
		\textbf{HRQE\_Decompose}                      & \multicolumn{1}{c}{$1.44\cdot10^{5}$}      & \multicolumn{1}{c}{$1.22\cdot10^{5}$}           &  $1.26\cdot10^{5}$      \\ \cline{1-4}
		\textbf{HRQE (ours)}                    & \multicolumn{1}{c}{$\textbf{2.97}\cdot\textbf{10}^{\textbf{2}}$}      & \multicolumn{1}{c}{$\textbf{1.35}\cdot\textbf{10}^{\textbf{3}}$}           & $\textbf{3.55}\cdot\textbf{10}^{\textbf{2}}$     \\ \hline
	\end{tabular}
	\vspace{-15pt}
\end{table}
%{\color{blue}
%\subsubsection{Decomposition:}
%}

%\begin{figure}[h] 
%	
%	\centering 	
%	%	\hfill
%	%\vspace{-10pt}
%	\subfloat[WD50K]	{\centering\includegraphics[width = 0.5\linewidth]{figs/overall/wd50k_overall2_fact_size_boxplot.pdf}}
%	\hfill
%	\subfloat[wikipeople]	{\centering\includegraphics[width = 0.5\linewidth]{figs/overall/wikipeople_overall2_fact_size_boxplot.pdf}}
%	%	\hfill
%	%	\subfloat[JF17K]	{\centering\includegraphics[width = 0.45\linewidth]{figs/overall/jf17k_overall2_fact_size_boxplot.pdf}}
%	
%	%	\label{ssec:tendency:tendecy}
%	%	\label{ssec:tendency:tendecy}
%	%\vspace{-10pt}
%	\caption{Q-Error boxplots grouping by query fact sizes over WD50K and Wikipeople}
%	%\vspace{-10pt}
%	\label{fig:factsize}
%\end{figure}

% HRQE-Cat changes the qualifier aggregation mechanism to concat instead of default weighted-sum. 

%The effectiveness comparison for qualifier completion mechansim is included in Section~\ref{ssec:para}.
%From Table \ref{tab:mainexp}, HRQE-Agg still achieves promising performance which proved the qualifier information is necessary in cardinality estimation for hyper-relational queries.
%However, it does not outperform HRQE as the qualifier completor is of vital importance to complete the missed qualifiers in query, where the lightweight generative model (CVAE) is suitable to model the conditional distribution between qualifiers and main triple.

\subsection{Parameter Sensitivity} \label{ssec:para}
In this section, we mainly study the performance of HRQE under different number of GNN layers $L$ and different value of $\lambda$ in Equation~\eqref{eq:lambda} by boxplots in  Figure~\ref{fig:layer} and Figure~\ref{fig:lambda}. 
%We use boxplot to compare the q-error under different hyperparameter settings.
\begin{figure*}[h] 
	
	\centering\includegraphics[width=\linewidth]{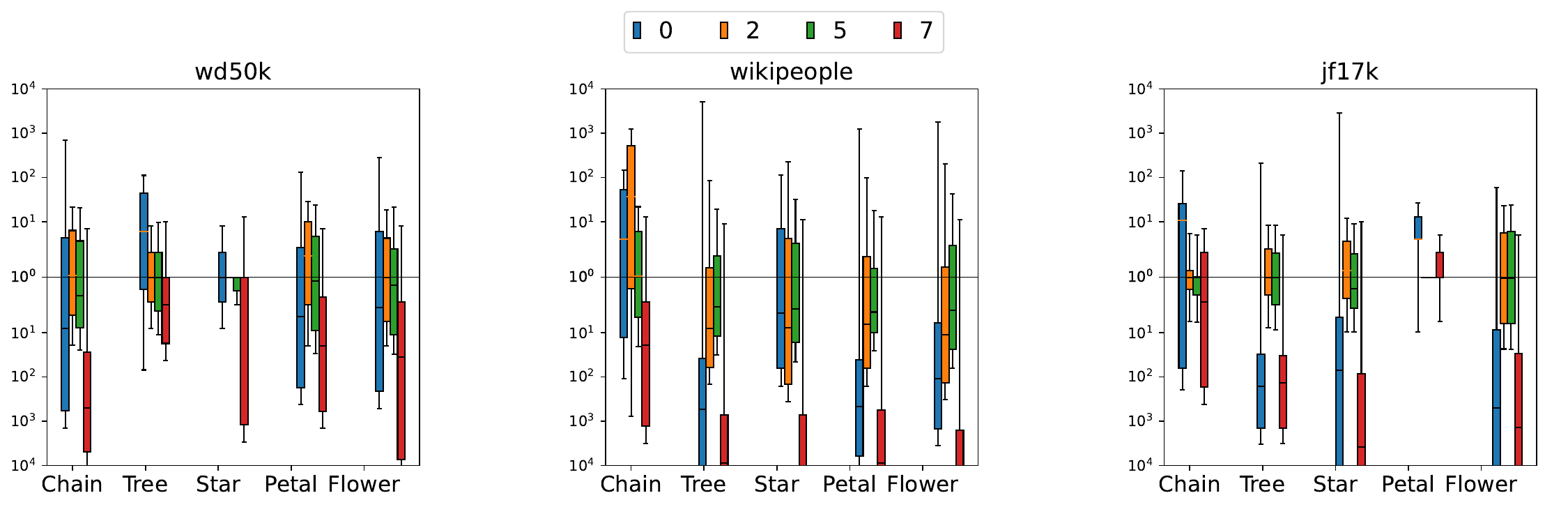}
	
	%	\label{ssec:tendency:tendecy}
	%	\label{ssec:tendency:tendecy}
%	\vspace{-10pt}
	\caption{Q-Error boxplots of varying GNN layer number over three datasets}
	\vspace{-10pt}
	\label{fig:layer}
\end{figure*}
\begin{figure*}[h] 
	
	\centering\includegraphics[width=\linewidth]{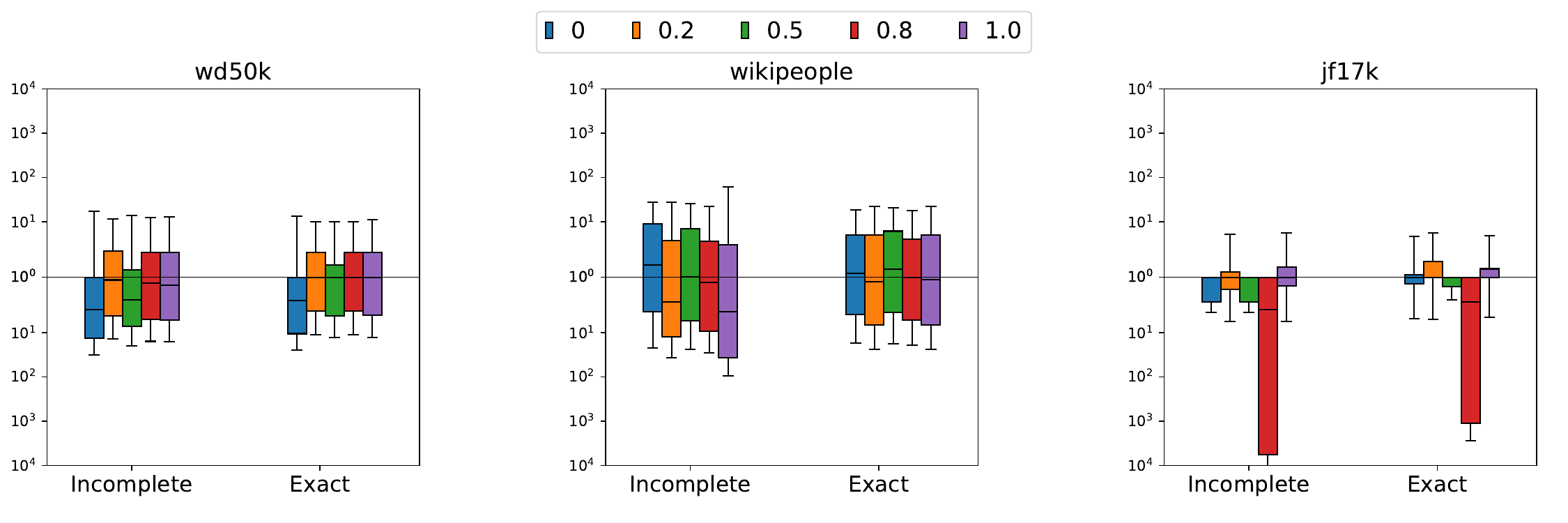}
	
	%	\label{ssec:tendency:tendecy}
	%	\label{ssec:tendency:tendecy}
%	\vspace{-10pt}
	\caption{Q-Error boxplots of varying $\lambda$ over three datasets}
	\vspace{-10pt}
	\label{fig:lambda}
\end{figure*}

\subsubsection{The number of GNN layers}

As for GNN layer number $L$, we compare the performance of query encoder on five query patterns varying the number of layers within $\{0, 2, 5, 7\}$ with other parameters maintaining the best setting. Here 0-layer refers to directly use the initialized embedding as the final node representation.
The boxplots for 0 layer indicates the necessity of message-passing.
As shown in Figure~\ref{fig:layer},
with the increase of layer numbers, the boxplots become narrower first, 
indicating increasing the number of layers is an effective way to handle diverse query patterns. 
Besides, the boxplots on 7 layers become wider compared to smaller layer number over three HKGs, which indicate that the receptive fields for nodes are not simply the larger the better.
It is because enlarging the receptive fields without regularization may cause oversmoothing problem~\cite{oversmooth}. Besides, the enlarged number of layers will also increase the difficulty of model training.
Thus, we need to keep a medium value of $L$. 
\subsubsection{The weight $\lambda$ of completed qualifiers} \label{ssec:qualcomple}

$\lambda$ is hyper-parameter to control the weight of qualifier completed query embedding in Equation~\eqref{eq:lambda}. We vary the value among $\{0, 0.2, 0.5, 0.8, 1.0\}$.
We introduce a new grouping criteria that refers to queries with incomplete qualifiers as \textit{incomplete}, otherwise they are called \textit{exact}. 
As shown in Figure~\ref{fig:lambda}, the encoder reached the performance peak at around $0.0$ over wikipeople queryset since it has the narrowest bar on both incomplete and exact queries.  
However, in queryset upon WD50K, the q-error bar is more centralized at 0 and shorter between 0.8 and 1.0, indicating the encoder achieves its best performance in range of [0.8, 1.0] over WD50K.
According Table~\ref{tab:stat}, only $2.7\%$ facts in wikipeople have qualifiers. The completed embedding might be a sort of noise. On the contrary, facts with qualifiers account for $13.6\%$ and $45.9\%$ in WD50K and JF17K respectively, which means that the fact with qualifiers are dominates on WD50K. Besides, the number of qualifier pairs within one fact is larger on WD50K, causing the higher demand for qualifier completion. The higher $\lambda$ value indicates that the proposed qualifier completer could be an effective technique to complete the missing qualifiers from existing fact pattern atoms on qualifier intensive HKGs. 
%Though WD50K have less qualifiered facts, the number of facts with large number of qualifier pairs is larger than that of JF17K, which increases the chance of incomplete qualifiers.

%\begin{figure*}[h]
%	\centering 	
%	%	\hfill
%	%\vspace{-10pt}
%	\subfloat[WD50K]	{\centering\includegraphics[width = 0.33\linewidth]{figs/lambda/wd50k_lambda_incomplete_boxplot.pdf}}
%	\hfill
%	\subfloat[wikipeople]	{\centering\includegraphics[width = 0.33\linewidth]{figs/lambda/wikipeople_lambda_incomplete_boxplot.pdf}}
%		\hfill
%		\subfloat[JF17K]	{\centering\includegraphics[width = 0.33\linewidth]{figs/lambda/jf17k_lambda_incomplete_boxplot.pdf}}
%	
%	%	\label{ssec:tendency:tendecy}
%	%\vspace{-10pt}
%	
%	%	\label{ssec:tendency:tendecy}
%	\caption{Q-Error boxplots of varying $\lambda$ over three datasets}
%	%\vspace{-15pt}
%	\label{fig:lambda}
%	
%\end{figure*}

{\color{black}
	\subsection{Case Study} \label{exp:case}
	
	In this section, we mainly study the performance of HRQE and several baselines under different type of queries. We split queries over WD50K~\cite{StarE}, Wikipeople~\cite{wikipeople} and JF17K~\cite{JF17K} from three dimensions, including query pattern, fact sizes and query with/without incomplete qualifiers. We plot the inference q-error for each dimension by boxplots in Figure~\ref{fig:topoexp}, Figure~\ref{fig:factsize} and Figure~\ref{fig:incompleteexp} respectively. 
	
	\begin{figure*}[t]
		%	\centering 	
		%	%\vspace{-10pt}
		%	% 	\hfill
		%	\subfloat[WD50K]	{\centering\includegraphics[width = 0.33\linewidth]{figs/overall/wd50k_overall2_shape_boxplot.pdf}}
		%	\hfill
		%	\subfloat[wikipeople]	{\centering\includegraphics[width = 0.33\linewidth]{figs/overall/wikipeople_overall2_shape_boxplot.pdf}}
		%	\hfill
		%	\subfloat[JF17K]	{\centering\includegraphics[width = 0.33\linewidth]{figs/overall/jf17k_overall2_shape_boxplot.pdf}}
		%	\subfloat[WD50K]	{\centering\includegraphics[width = 0.33\linewidth]{figs/overall/wd50k_overall2_shape_boxplot.pdf}}
		%	\label{ssec:tendency:tendecy}
		\centering\includegraphics[width=\linewidth]{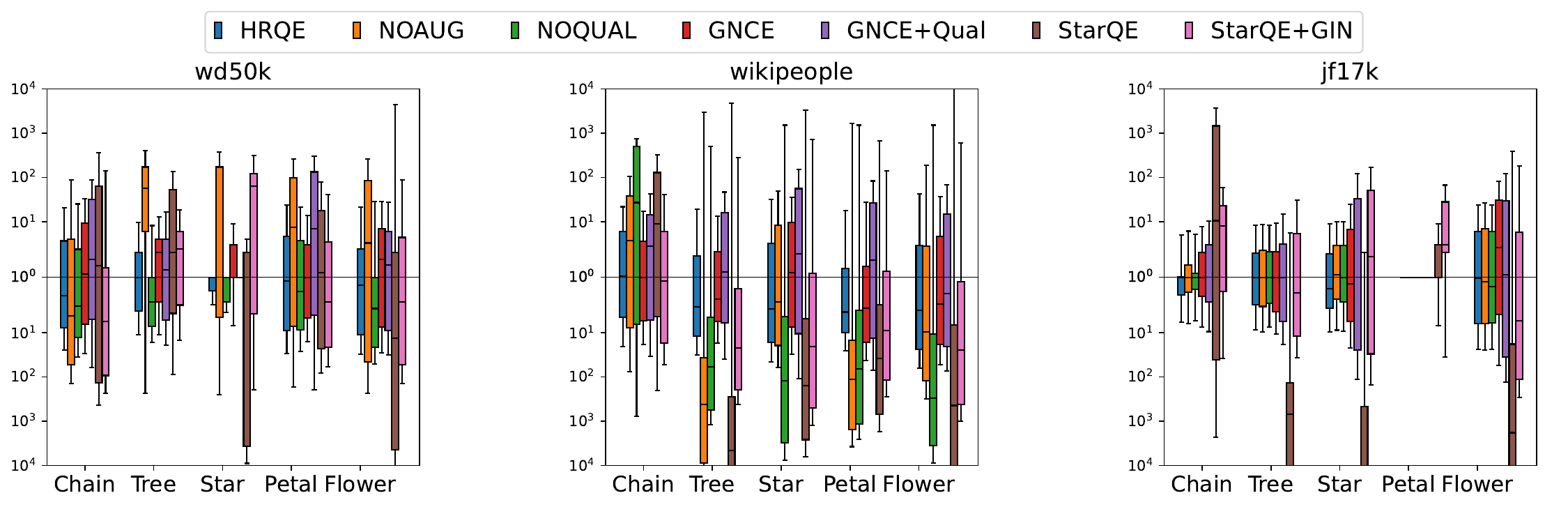}
		\vspace{-10pt}
		%	\label{ssec:tendency:tendecy}
		\caption{Q-Error boxplots grouping by query pattern over three datasets}
		\vspace{-10pt}
		\label{fig:topoexp}
		
	\end{figure*}
	
	\begin{figure*}[h] 
		
		\centering\includegraphics[width=\linewidth]{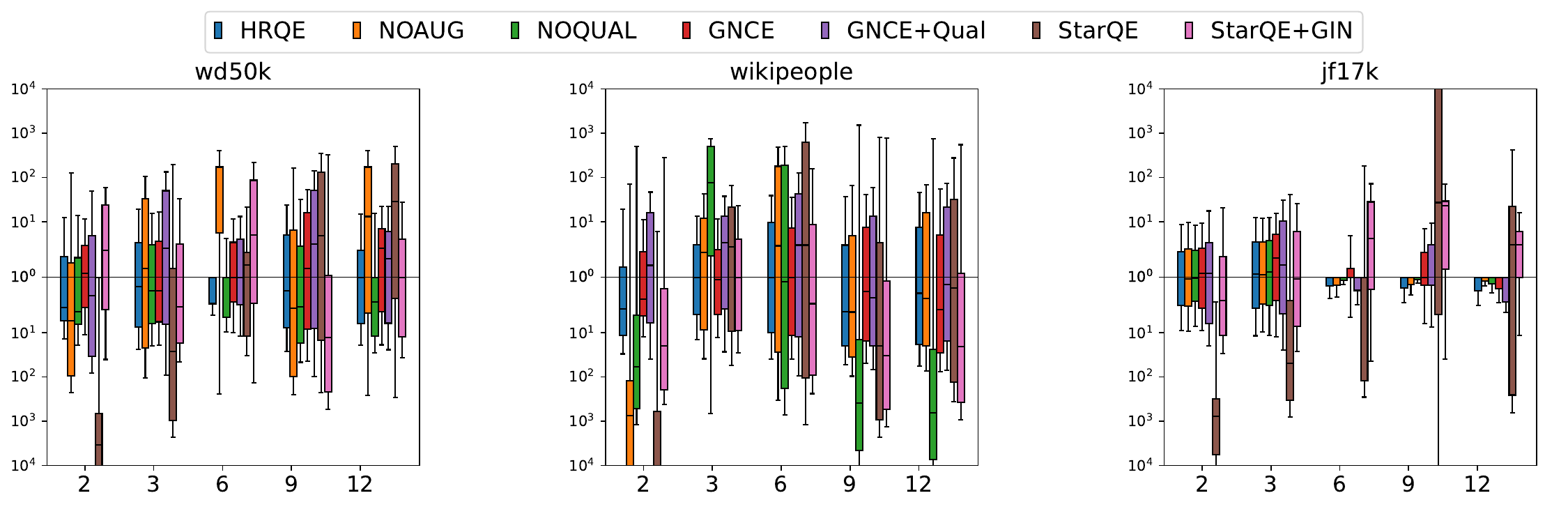}
		
		%	\label{ssec:tendency:tendecy}
		%	\label{ssec:tendency:tendecy}
		\vspace{-10pt}
		\caption{Q-Error boxplots grouping by query fact sizes over three datasets}
		\vspace{-10pt}
		\label{fig:factsize}
	\end{figure*}

	\begin{figure*}[h]
		\centering\includegraphics[width=\linewidth]{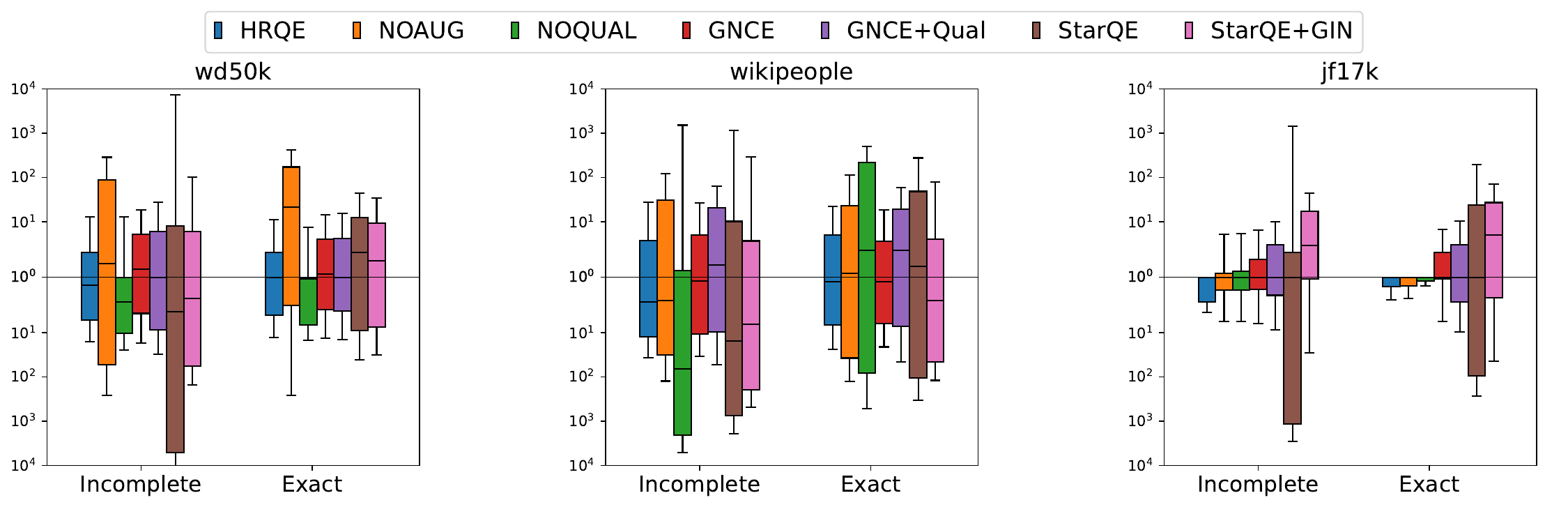}
		
		\vspace{-10pt}
		%	\label{ssec:tendency:tendecy}
		\caption{Q-Error boxplots grouping by incomplete queries over three datasets}
		\vspace{-10pt}
		\label{fig:incompleteexp}
		
	\end{figure*}
	
	\subsubsection{Query Patterns} 
	
	We group the queries into five patterns, Chain, Tree, Star, Petal and Flower, introduced in Section~\ref{ssec:query_HKGs} and compare HRQE with other competitive baselines over the five patterns in Figure~\ref{fig:topoexp}. 
	Generally, petal and flower queries are more complex patterns that have larger q-error than other three acyclic patterns. 
	Thanks to the multiple message passing layers design which enlarges the receptive fields of nodes and improves the capability for complex query patterns. Among all baselines and ablation variants, HRQE has the narrowest boxes in all five patterns over three datasets, indicating it can fit multiple types of query patterns.
	
	\subsubsection{Fact Sizes} 
	
	Figure~\ref{fig:factsize} illustrates the effectiveness of HRQE and baselines under different number of facts.
	HRQE maintains its leading effectiveness across different fact sizes.
	Besides, all methods perform better on queries with less facts like 2 and 3 while the effectiveness degrades as fact size increasing. Compared to other models, HRQE is more robust when fact size increases because our proposed data augmentation strategy trains HRQE over more diverse query set and let the model generalizes well to various fact sizes.
	
	\subsubsection{Query with incomplete qualifiers} 
	
	To verify the effectiveness of qualifier completer in Section~\ref{ssec:qualifier_completion}, we also separate the queries into two groups where one group has incomplete qualifiers while the other does not, namely incomplete and exact. According to Figure~\ref{fig:incompleteexp}, incomplete queries have larger estimation error compared to exact group over WD50K, illustrating the necessity of qualifier completion.
	According to Table~\ref{tab:stat}, over 45.9\% of facts in JF17K contain qualifiers, but the number of qualifier pairs within one fact is fewer than in WD50K. This indicates the qualifier pair distribution in JF17K is simpler than that in WD50K. The situation is similar on that in Wikipeople where only 2.3\% facts have qualifiers.
	Thanks to parameter $\lambda$ in Equation~\eqref{eq:lambda}, we can manually control the effects of qualifier completion mechanism so that HRQE still outperforms all baselines over both incomplete and exact queries.
}

\begin{figure}[t]
	\centering 	
	%\vspace{-15pt}
	\includegraphics[width=0.8\linewidth]{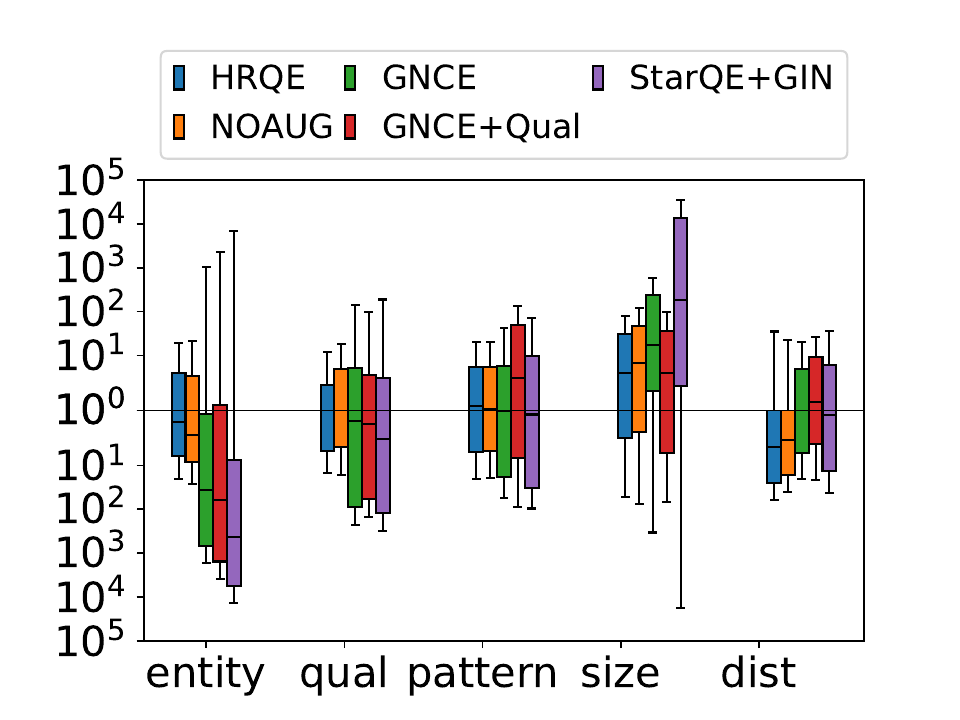}	
	%	{\centering\includegraphics[width = 0.32\linewidth]{}}
	%	\label{ssec:tendency:tendecy}
	
	\vspace{-10pt}
	%	\label{ssec:tendency:tendecy}
	\caption{Inductive evaluation grouping by query patterns, fact sizes, qualifiers and entities over WD50K}
	\vspace{-20pt}
	\label{fig:inductive}
	
\end{figure}
\subsection{Inductive Evaluation} \label{exp:inductive}

	%estimator, stare encoder, augmentation, remove qualifier
	In this subsection, we test the effectiveness of HRQE in inductive cases, where test queries contain elements not seen in train queries. 
	The inductive ability is a major concern for CE model in production environments~\cite{GNCE}, as it is costly to retrain or update the CE model.
%	, leading to requirement that CE model is supposed to generalize well to unseen type of queries given a proper training set. 
	To build such train/test queries, we re-split queryset in WD50K by four criteria, query patterns, fact sizes, whether contain qualifiers (qualifiers for short) and whether contain bounded entities (entities for short). For each inductive cases, we select 7000 training queries and 3500 test queries.
	
	We compare HRQE with GNCE, GNCE+Qual and StarQE+GIN, the competitive baseline models with competitive overall performance. Besides, we also select HRQE-NoAug as a baseline since the relative cardinality based data augmentation strategy should be effective to help model generalize to unseen queries.
	
	%From Table \ref{tab:mainexp}, HRQE-Agg still achieves promising performance which proved the qualifier information is necessary in cardinality estimation for hyper-relational queries.
	%However, it does not outperform HRQE as the qualifier completor is of vital importance to complete the missed qualifiers in query, where the lightweight generative model (CVAE) is suitable to model the conditional distribution between qualifiers and main triple.

	\subsubsection{Inductive query patterns}  we randomly select 7000 acyclic queries in WD50K as training set and select 3500 cyclic queries as test set.
	The group \textit{pattern} in Figure~\ref{fig:inductive} shows that all the methods suffer performance deterioration compared to overall case due to the difference between training/test query patterns. HRQE outperforms baselines in inductive cyclic queries.

%Most of previous works~\cite{LSS, GNCE, ALLEY, GCARE, fastest} have proved that cyclic query patterns i.e. petal and flower are rarer and more complex compared to acyclic query patterns i.e. chain, tree and star. To test the inductive ability in regard to query patterns,	
	\subsubsection{Inductive fact sizes} We randomly select 7000 queries with $\leq 6$ facts in WD50K as training set and select 3500 queries $\geq 9$ as test set.
	 According to \textit{size} in Figure~\ref{fig:inductive}, though all the methods suffer performance deterioration compared to overall case. HRQE outperforms GNCE, GNCE+Qual and StarQE+GIN in larger test queries, proving that the proposed model, including qualifier-aware
	  message passing and more GNN layers helps to generalize better to larger queries. Data augmentation improves the inductive ability as well according to comparison between HRQE and its no augmentation version. 
	 
%	  Most of previous works~\cite{ALLEY, GCARE, fastest} have proved that the larger fact size of query leads to increasingly higher q-error. To test the inductive ability in regard to fact size of query, 
	
%	{\color{blue}
	\subsubsection{Inductive qualifier queries} We randomly select 7000 queries without qualifiers in WD50K as training set and select 3500 queries with qualifiers as test set.
	The \textit{qual} in Figure~\ref{fig:inductive} shows that all the methods has increased q-error compared to overall cases. Thanks to qualifier add/removal data augmentation strategy, HRQE outperforms HRQE-NoAug, which also outperforms other baseline models, proving that HRQE is more suitable for handling qualifiers.
	{\color{black}
	\subsubsection{Inductive qualifiers distribution} To study the potential performance variance between qualifier pairs distribution in training set and test set, we randomly select test queries where the qualifier pairs in training set are not presenting in test set.
	The \textit{dist} in Figure~\ref{fig:inductive} shows that all the methods has increased q-error compared to overall cases. Thanks to the proposed qualifier completer, HRQE outperforms all baselines, proving that our proposed model is more stable in different qualifier pairs distribution.

}
	\subsubsection{Inductive entities} To prevent HRQE simply memorizing the entity embedding, we randomly select 7000 queries with bounded entities in WD50K as training set and select 3500 queries with no bounded entities as test set.
	According to \textit{entity} in Figure~\ref{fig:inductive}, HRQE and HRQE-NoAug outperforms other baseline models, proving that our model suffer less from the potential entity memorization problem. HRQE is more robust when estimating queries with entities not seen in training.
	
		\begin{table}[t] 
		\centering
		\vspace{-5pt}

		\caption{Effectiveness (Mean q-Errors) for learning-based models over inductive large queries. 
			The \textbf{bold number} and the \underline{underline number} indicate the best and the second best performance, respectively.	
		}
				\vspace{-5pt}
		\label{tab:inductivelarge}
		\begin{tabular}{c|c|c|c}
			\hline
			\multirow{1}{*}{\textbf{Model}}  & \multicolumn{1}{c}{\textbf{Mean q-Errors}}  & \multicolumn{1}{c}{\textbf{Inference Time (s)}}                         \\ \cline{1-3} 
			
			\textbf{StarQE+GIN}                     & \multicolumn{1}{c}{$4.09\cdot10^{4}$}  & \multicolumn{1}{c}{$\textbf{6.97}\cdot\textbf{10}^{\textbf{-3}}$}    \\ 
			\textbf{GNCE}                       & \multicolumn{1}{c}{$5.61\cdot10^{3}$}   & \multicolumn{1}{c}{$3.33\cdot10^{-2}$}      \\ 
			\textbf{GNCE+Qual}                       & \multicolumn{1}{c}{$5.76\cdot10^{3}$}  & \multicolumn{1}{c}{\underline{$7.17\cdot10^{-3}$}}         \\ 
			\textbf{HRQE\_NoAug}                   & \multicolumn{1}{c}{\underline{$4.93\cdot10^{3}$}}     & \multicolumn{1}{c}{$1.51\cdot10^{-2}$}    \\  \hline
			\textbf{HRQE (ours)}                    & \multicolumn{1}{c}{$\textbf{4.27}\cdot\textbf{10}^{\textbf{3}}$} & \multicolumn{1}{c}{$1.86\cdot10^{-2}$}        \\ \hline
		\end{tabular}
		\vspace{-10pt}
	\end{table}

	\subsubsection{Inductive large queries} To evaluate the performance of HRQE over large hyper-relational query graphs, we run Algorithm~\ref{alg:querygeneration} to generate 300 query graphs over WD50K where each query graph has more than 20 nodes and 30 edges. We train HRQE and several comparable baselines over default training query set of WD50K and verify the average q-error and inference time in Table~\ref{tab:inductivelarge}. 
	As shown in table, all methods suffer from performance downgrade since the largest training query only has 12 nodes connected via 16 edges. HRQE still outperforms all the baselines, indicating the better generalization ability to unseen queries. Furthermore, compared to results in Table~\ref{tab:mainexp}, HRQE has good scalability to large query in terms of inference time, as inference complexity is linear to the number of nodes in query. 

%}

\section{Conclusion}\label{sec:conclusion}
In this paper, we comprehensively investigate the cardinality estimation of queries on hyper-relational knowledge graphs.
We first constructed diverse and unbiased hyper-relational query sets over three popular HKGs.  Additionally, we proposed a novel qualifier-aware graph neural network (GNN) model with simple and effective data augmentation, which effectively incorporates qualifier information and adaptively combines outputs from multiple GNN layers for accurate cardinality prediction. 
Our experiments illustrate that the proposed hyper-relational query encoder outperforms state-of-the-art CE methods across the three HKGs on our diverse and unbiased benchmark.
%	However, our work only focuses on conjunctive HKG query which is only a subset among diverse real-world logical operators for HKG queries. 
	In the future, we will extend the current model to support multiple logical operators, such as  negation operation~\cite{NQE} on cardinality estimation. 

%In the future, we will directly apply our proposed model on the downstream tasks, such as query optimization and property distribution prediction, to comprehensively evaluate its effectiveness.
%In the future, we will focus on cardinality estimation task

\bibliographystyle{ACM-Reference-Format}
\bibliography{sample-base.bib}
 \clearpage
\end{document}